\documentclass[10pt,twocolumn,letterpaper]{article}

\usepackage[pagenumbers]{cvpr} 

\usepackage{graphicx}
\usepackage{amsmath}
\usepackage{amssymb}
\usepackage{booktabs}
\usepackage{soul}

\usepackage[pagebackref,breaklinks,colorlinks]{hyperref}

\usepackage[capitalize]{cleveref}
\crefname{section}{Sec.}{Secs.}
\Crefname{section}{Section}{Sections}
\Crefname{table}{Table}{Tables}
\crefname{table}{Tab.}{Tabs.}

\makeatletter
\@namedef{ver@everyshi.sty}{}
\makeatother

\usepackage{amsmath}
\usepackage{amsfonts}
\usepackage{amssymb}
\usepackage{wrapfig}
\usepackage{subcaption}
\usepackage{multirow}
 \usepackage{mathtools} 

\usepackage{verbatim}

\usepackage{anyfontsize}

\usepackage{microtype}
\usepackage{graphicx}
\usepackage{booktabs} 
\usepackage{multirow}
\usepackage{amsmath,amssymb}
\usepackage{booktabs}
\usepackage{caption,subcaption}

\usepackage{xcolor}
\definecolor{mygreen}{HTML}{3cb44b}
\definecolor{skyblue}{HTML}{beffff}
\definecolor{lightgreen}{HTML}{90ee90}

\usepackage{tcolorbox}
\usepackage{enumitem}
\setitemize{itemsep=10pt,topsep=0pt,parsep=0pt,partopsep=0pt}
\usepackage{adjustbox}

\newcommand{\RN}[1]{%
	\textup{\lowercase\expandafter{\it \romannumeral#1}}%
}
\usepackage{tabu}

\usepackage{amsmath}

\newcommand{\beq}{\vspace{0mm}\begin{equation}}
\newcommand{\eeq}{\vspace{0mm}\end{equation}}
\newcommand{\beqs}{\vspace{0mm}\begin{eqnarray}}
\newcommand{\eeqs}{\vspace{0mm}\end{eqnarray}}
\newcommand{\barr}{\begin{array}}
\newcommand{\earr}{\end{array}}

\newcommand{\Pmat}{{\bf P}}

\newcommand{\Umat}[0]{{{\bf U}}}

\newcommand{\cv}[0]{{\boldsymbol{c}}}

\newcommand{\ev}[0]{{\boldsymbol{e}}\xspace}

\newcommand{\hv}[0]{{\boldsymbol{h}}}

\newcommand{\kv}[0]{{\boldsymbol{k}}\xspace}
\newcommand{\lv}[0]{{\boldsymbol{l}}}

\newcommand{\pv}[0]{{\boldsymbol{p}}}

\newcommand{\uv}{\boldsymbol{u}}
\newcommand{\vv}{\boldsymbol{v}}

\newcommand{\xv}{\boldsymbol{x}}
\newcommand{\yv}{\boldsymbol{y}}
\newcommand{\zv}{\boldsymbol{z}}

\newcommand{\epsilonv}{\boldsymbol{\epsilon}}

\newcommand{\thetav}{\boldsymbol{\theta}}

\usepackage{color, colortbl}
\definecolor{Gray}{gray}{0.93}

\usepackage{lipsum}

\newcommand\blfootnote[1]{%
  \begingroup
  \renewcommand\thefootnote{}\footnote{#1}%
  \addtocounter{footnote}{-1}%
  \endgroup
}

\usepackage{xcolor,amsmath}
\usepackage[linesnumbered,ruled,vlined]{algorithm2e}
\DontPrintSemicolon

\usepackage{color, colortbl}

\definecolor{emerald}{rgb}{0.31, 0.78, 0.37}

\newcommand{\MyColorBox}[2][red]%
{%
    \settowidth{\Width}{#2}%
    \colorbox{#1}%
    {%
        \raisebox{-\DepthReference}%
        {%
                \parbox[b][\HeightReference+\DepthReference][c]{\Width}{\centering#2}%
        }%
    }%
}

\definecolor{codegray}{gray}{0.9}

\SetKwComment{Comment}{\color{green!50!black}\# }{}

\SetKwProg{Function}{def}{:}{}

\SetKwProg{For}{for}{:}{}
\SetKwProg{If}{if}{:}{}

\def\cvprPaperID{3041} 
\def\confName{CVPR}
\def\confYear{2023}

\newcommand{\shortname}{\textsc{Gligen}}
\newcommand{\longname}{\textbf{G}rounded-\textbf{L}anguage-to-\textbf{I}mage \textbf{Gen}eration}

\usepackage[accsupp]{axessibility}  

\begin{document}

\title{\shortname{}: Open-Set Grounded Text-to-Image Generation}


\author{
Yuheng Li$^{1\S}$, ~Haotian Liu$^{1\S}$,  ~Qingyang Wu$^{2}$, ~Fangzhou Mu$^{1}$,  ~Jianwei Yang$^{3}$, ~Jianfeng Gao$^{3}$, \\
Chunyuan Li$^{{3}{\P}}$, ~Yong Jae Lee$^{1\P}$
\and
\and
{
\small
\textsuperscript{1}\textbf{University of Wisconsin-Madison} \;
\textsuperscript{2}\textbf{Columbia University} \;
\textsuperscript{3}\textbf{Microsoft} \;
}
\and
\url{https://gligen.github.io/}
}

\twocolumn[{%
	\maketitle
	\renewcommand\twocolumn[1][]{#1}%
	\begin{center}
		\centering
            \includegraphics[width=0.94\textwidth]{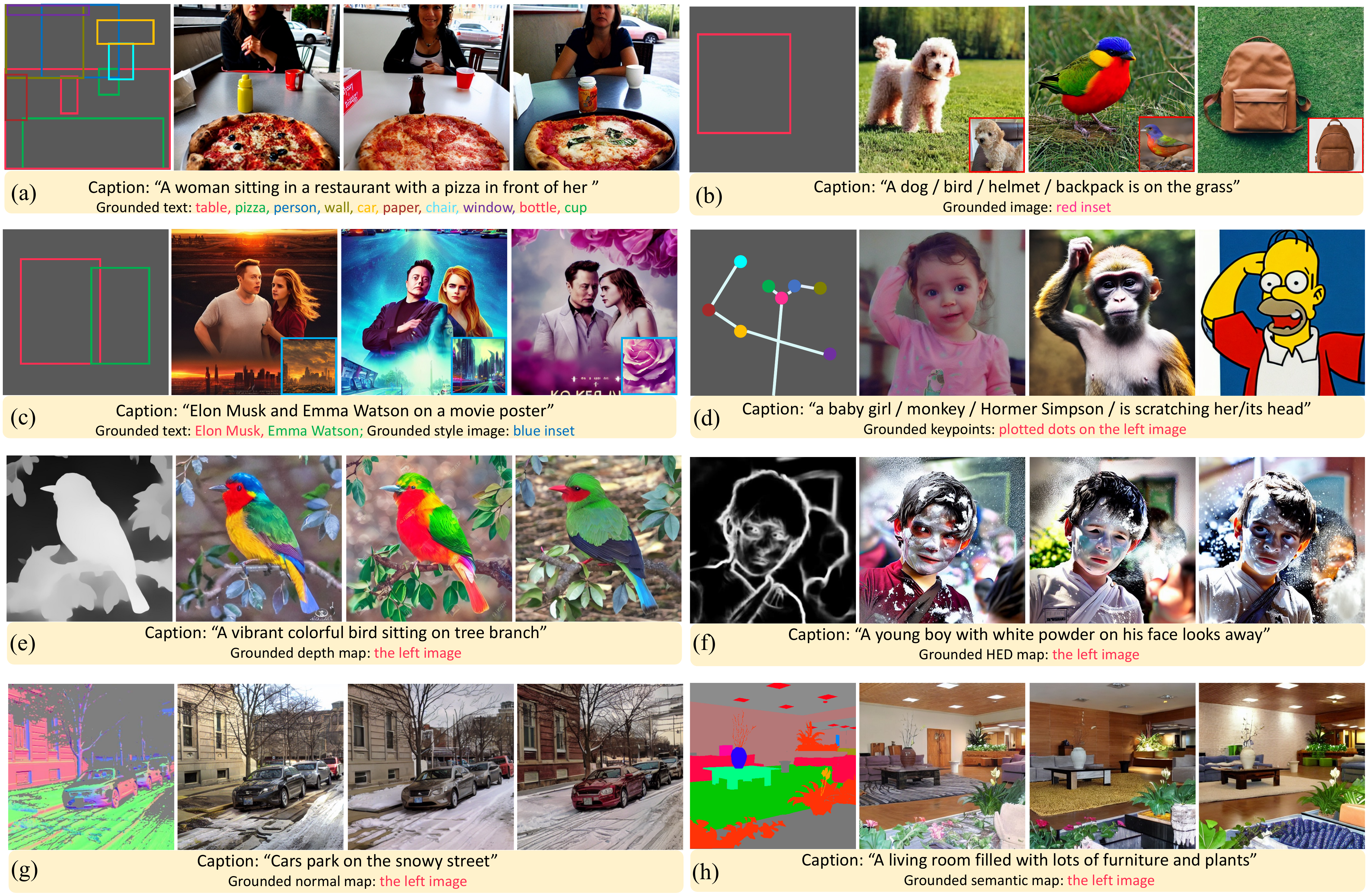}
		\captionof{figure}{\shortname{} enables versatile grounding capabilities for a frozen text-to-image generation model, by feeding different grounding conditions. \shortname{} supports (a) text entity +  box, (b) image entity +  box, (c) image style and text + box, (d)  keypoints,  (e) depth map, (f) edge map, (g) normal map, and (h) semantic map. }
		\label{fig:teaser}
		\vspace{0.1in}
	\end{center}
}]
%

\begin{abstract}
\vspace{-10pt}
Large-scale text-to-image diffusion models have made amazing advances. However, the status quo is to use text input alone, which can impede controllability. In this work, we propose \shortname{}, \longname{}, a novel approach that builds upon and extends the functionality of existing pre-trained text-to-image diffusion models by enabling them to also be conditioned on grounding inputs. To preserve the vast concept knowledge of the pre-trained model, we freeze all of its weights and inject the grounding information into new trainable layers via a gated mechanism.  Our model achieves open-world grounded text2img generation with caption and bounding box condition inputs, and the grounding ability generalizes well to novel spatial configurations and concepts. \shortname{}'s zero-shot performance on COCO and LVIS outperforms existing supervised layout-to-image baselines by a large margin. 
\vspace{0pt}
\blfootnote{$\S$ Part of the work performed at Microsoft; ${\P}$~Co-senior authors }
\end{abstract}

\vspace{-25pt}
\section{Introduction}\label{sec:intro}
\vspace{-3pt}

Image generation research has witnessed huge advances in recent years.  Over the past couple of years, GANs~\cite{Goodfellow2014GenerativeAN} were the state-of-the-art, with their latent space and conditional inputs being well-studied for controllable manipulation~\cite{Pathak2016ContextEF,interfacegan} and generation~\cite{stylegan,spade,stylegan2,Lafite}. Text conditional autoregressive~\cite{DALLE,PARTI} and diffusion~\cite{DALLE2,Imagen} models have demonstrated astonishing image quality and concept coverage, due to their more stable learning objectives and large-scale training on web image-text paired data. These models have gained attention even among the general public due to their practical use cases (\eg, art design and creation).

Despite exciting progress, existing large-scale text-to-image generation models cannot be conditioned on other input modalities apart from text, and thus lack the ability to precisely localize concepts, use reference images, or other conditional inputs to control the generation process. The current input, \ie, natural language alone, restricts the way that information can be expressed.  For example, it is difficult to describe the precise location of an object using text, whereas bounding boxes / keypoints can easily achieve this, as shown in Figure~\ref{fig:teaser}.  While conditional diffusion models~\cite{diffusionbeatsgan,Saharia2022PaletteID,LDM} and GANs~\cite{Pathak2016ContextEF,contrasfill,Johnson2018ImageGF, twfa} that take in input modalities other than text for inpainting, layout2img generation, \etc, do exist, they rarely combine those inputs for controllable text2img generation.

Moreover, prior generative models---regardless of the generative model family---are usually independently trained on each task-specific dataset.
In contrast, in the recognition field, the long-standing paradigm has been to build recognition models~\cite{li2022elevater, liu2023react, zou2022xdecoder} by starting from a foundation model pretrained on large-scale image data~\cite{he2020momentum,bao2021beit,he2022masked} or image-text pairs~\cite{clip,li2021align,yuan2021florence}.
Since diffusion models have been trained on billions of image-text pairs~\cite{LDM}, a natural question is: \emph{Can we build upon existing pretrained diffusion models and endow them with new conditional input modalities?}
In this way, analogous to the recognition literature, we may be able to achieve better performance on other generation tasks due to the vast concept knowledge that the pretrained models have, while acquiring more controllability over existing text-to-image generation models.

With the above aims, we propose a method for providing new grounding conditional inputs to pretrained text-to-image diffusion models. As shown in Figure~\ref{fig:teaser}, we still retain the text caption as input, but also enable other input modalities such as bounding boxes for grounding concepts, grounding reference images, grounding part keypoints, etc. The key challenge is preserving the original vast concept knowledge in the pretrained model while learning to inject the new grounding information. To prevent knowledge forgetting, we propose to freeze the original model weights and add new trainable gated Transformer layers~\cite{Vaswani2017AttentionIA} that take in the new grounding input (\eg, bounding box). During training, we gradually fuse the new grounding information into the pretrained model using a gated mechanism~\cite{Alayrac2022FlamingoAV}. This design enables flexibility in the sampling process during generation for improved quality and controllability; for example, we show that using the full model (all layers) in the first half of the sampling steps and only using the original layers (without the gated Transformer layers) in the latter half can lead to generation results that accurately reflect the grounding conditions while also having high image quality.

In our experiments, we primarily study grounded text2img generation with bounding boxes, inspired by the recent scaling success of learning grounded language-image understanding models with boxes in GLIP~\cite{GLIP}.To enable our model to ground open-world vocabulary concepts~\cite{zareian2021open,GLIP,zhong2022regionclip,li2022elevater}, we use the same pre-trained text encoder (for encoding the caption) to encode each phrase associated with each grounded entity (\ie, one phrase per bounding box) and feed the encoded tokens into the newly inserted layers with their encoded location information. Due to the shared text space, we find that our model can generalize to unseen objects even when only trained on the COCO~\cite{coco} dataset. Its generalization on LVIS~\cite{lvis} outperforms a strong fully-supervised baseline by a large margin. To further improve our model's grounding ability, we unify the object detection and grounding data formats for training, following GLIP~\cite{GLIP}. With larger training data, our model's generalization is consistently improved.

\vspace{-10pt}
\paragraph{Contributions.} 1) We propose a new text2img generation method that endows new grounding controllability over existing text2img diffusion models. 2) By preserving the pre-trained weights and learning to gradually integrate the new localization layers, our model achieves open-world grounded text2img generation with bounding box inputs, \ie, synthesis of novel localized concepts unobserved in training. 3) Our model's zero-shot performance on layout2img tasks significantly outperforms the prior state-of-the-art, demonstrating the power of building upon large pretrained generative models for downstream tasks.

\section{Related Work}\label{sec:related}

\vspace{-3pt}
\paragraph{Large scale text-to-image generation models.} State-of-the-art models in this space are either autoregressive~\cite{DALLE,MAKEASCENE,PARTI,wu2022nuwa} or diffusion~\cite{GLIDE,DALLE2,LDM,Imagen,zhou2022shifted}. Among autoregressive models, DALL-E~\cite{DALLE} is one of the breakthrough works that demonstrates zero-shot abilities, while Parti~\cite{PARTI} demonstrates the feasibility of scaling up autoregressive models. Diffusion models have also shown very promising results. DALL-E 2~\cite{DALLE2} generates images from the CLIP~\cite{clip} image space, while Imagen~\cite{Imagen} finds the benefit of using pretrained language models. 
The concurrent Muse~\cite{chang2023muse} demonstrates that masked modeling can achieve SoTA-level generation performance with higher inference speed. 
However, all of these models usually only take a caption as the input, which can be difficult for conveying other information such as the precise location of an object. Make-A-Scene~\cite{MAKEASCENE} also incorporates semantic maps into its text-to-image generation, by training an encoder to tokenize semantic masks to condition the generation. However, it can only operate in a closed-set (of 158 categories), whereas our grounded entities can be open-world.  A concurrent work eDiff-I~\cite{ediffi} shows that by changing the attention map, one can generate objects that roughly follow a semantic map input. However, We believe our interface with boxes is simpler, and more importantly, our method allows other conditioning inputs such as keypoints, edge map, inference images, etc., which are hard to manipulate through attention.

\vspace{-10pt}
\paragraph{Image generation from layouts.} Given bounding boxes labeled with object categories, the task is to generate a corresponding image~\cite{Zhao2019ImageGF,lostgan,lostgan2,Sylvain2021ObjectCentricIG,Li2021ImageSF,Jahn2021HighResolutionCS,Yang2022ModelingIC}, which is the reverse task of object detection. Layout2Im~\cite{Zhao2019ImageGF} formulated the problem and combined a VAE object encoder, an LSTM~\cite{Hochreiter1997LongSM} object fuser, and an image decoder to generate the image, using global and object-level adversarial losses~\cite{Goodfellow2014GenerativeAN} to enforce realism and layout correspondence. LostGAN~\cite{lostgan,lostgan2} generates a mask representation which is used to normalize features, taking inspiration from StyleGAN~\cite{Karras2019ASG}. LAMA~\cite{Li2021ImageSF} improves the intermediate mask quality for better image quality. Transformer~\cite{NIPS2017_3f5ee243} based methods~\cite{Jahn2021HighResolutionCS,Yang2022ModelingIC} have also been explored. Critically, existing layout2image methods are closed-set, \ie, they can only generate limited localized visual concepts observed in the training set such as the 80 categories in COCO. In contrast, our method represents the first work for \emph{open-set} grounded image generation. A concurrent work ReCo~\cite{Reco} also demonstrates open-set abilities by building upon a pretraned Stable Diffusion model~\cite{LDM}. However, it finetunes the original model weights, which has the potential to lead to knowledge forgetting. Furthermore, it only demonstrates box grounding results whereas we show results on more modalities as shown in the Figure~\ref{fig:teaser}. 

\vspace{-10pt}
\paragraph{Other conditional image generation.} For GANs, various conditioning information have been explored; \eg, text~\cite{attngan,dfgan,zhou2022lafite2}, box~\cite{Zhao2019ImageGF,lostgan,lostgan2}, semantic masks~\cite{spade,collagegan}, images~\cite{cyclegan,stargan,mixnmatch}. For diffusion models, LDM~\cite{LDM} proposes a unified approach for conditional generation by injecting the condition via cross-attention layers. Palette~\cite{Saharia2022PaletteID} performs image-to-image tasks using diffusion models. These models are usually trained from scratch independently. In our work, we investigate how to build upon existing models pretrained on large-scale web data, to enable new open-set grounded image generation capabilities in a cost-effective manner.  

\section{Preliminaries on Latent Diffusion Models}

Diffusion-based methods are one of the most effective model families for text2image tasks, among which latent diffusion model (LDM)~\cite{LDM} and its successor Stable Diffusion are the most powerful models publicly available to the research community. 
To reduce the computational costs of vanilla diffusion model training, LDM proceeds in two stages. The first stage learns a bidirectional mapping network to obtain the latent representation $\zv$ of the image $\xv$.  The second stage trains a diffusion model on the latent $\zv$. Since the first stage model produces a fixed bidirectional mapping between $\xv$ and $\zv$, from hereon, we focus on the latent generation space of LDM for simplicity. 

\paragraph{Training Objective.}
Starting from noise $\zv_T$, the model gradually produces less noisy samples $\zv_{T-1}, \zv_{T-2}, \cdots, \zv_0$, conditioned on caption $\cv$ at every time step $t$. To learn such a model $f_{\thetav}$ parameterized by $\thetav$, for each step, the LDM training objective solves the denoising problem on latent representations $\zv$ of the image $\xv$: 
\begin{align}\label{eq:ldm_loss}
\min_{\thetav} \mathcal{L}_{\text{LDM}} = \mathbb{E}_{\zv, \epsilonv \sim  \mathcal{N}(\mathbf{0}, \mathbf{I}), t} \big[ \|  \epsilonv -   f_{\thetav}(\zv_t, t, \cv) \|^2_2 \big],
\end{align}
where $t$ is uniformly sampled from time steps $\{1, \cdots, T\}$, $\zv_t$ is the step-$t$ noisy variant of input $\zv$, and $ f_{\thetav} (*, t, \cv)$ is the $(t, \cv)$-conditioned denoising autoencoder.

\paragraph{Network Architecture.} The core of the network architecture is how to encode the conditions, based on which a cleaner version of $\zv$ is produced. 
$(i)$ {\it Denoising Autoencoder}.
$ f_{\thetav} (*, t, \cv)$ is
implemented via UNet~\cite{unet}. It takes in a noisy latent $\zv$, as well as information from time step $t$ and condition $\cv$. It consists of a series of ResNet~\cite{resnet} and Transformer~\cite{Vaswani2017AttentionIA} blocks. 
$(ii)$ {\it Condition Encoding}. In the original LDM, a BERT-like~\cite{bert} network is trained from scratch to encode each caption into a sequence of text embeddings, $f_{\text{text}}(\cv)$, which is fed into~\eqref{eq:ldm_loss} to replace $\cv$. The caption feature is encoded via a fixed CLIP~\cite{clip} text encoder in Stable Diffusion.   
Time $t$ is first mapped to time embedding $\phi(t)$, then injected into the UNet.
The caption feature is used in a cross attention layer within each Transformer block. The model learns to predict the noise, following \eqref{eq:ldm_loss}. 

With large-scale training, the model $ f_{\thetav} (*, t, \cv)$ is well trained to denoise $\zv$ based on the caption information only. Though impressive language-to-image generation results have been shown with LDM by pretraining on internet-scale data, it remains challenging to synthesize images where additional grounding input can be instructed, and is thus the focus of our paper.

\section{Open-set Grounded Image Generation}\label{sec:approach}

\begin{figure}[t!]
    \centering
    \includegraphics[width=0.45\textwidth]{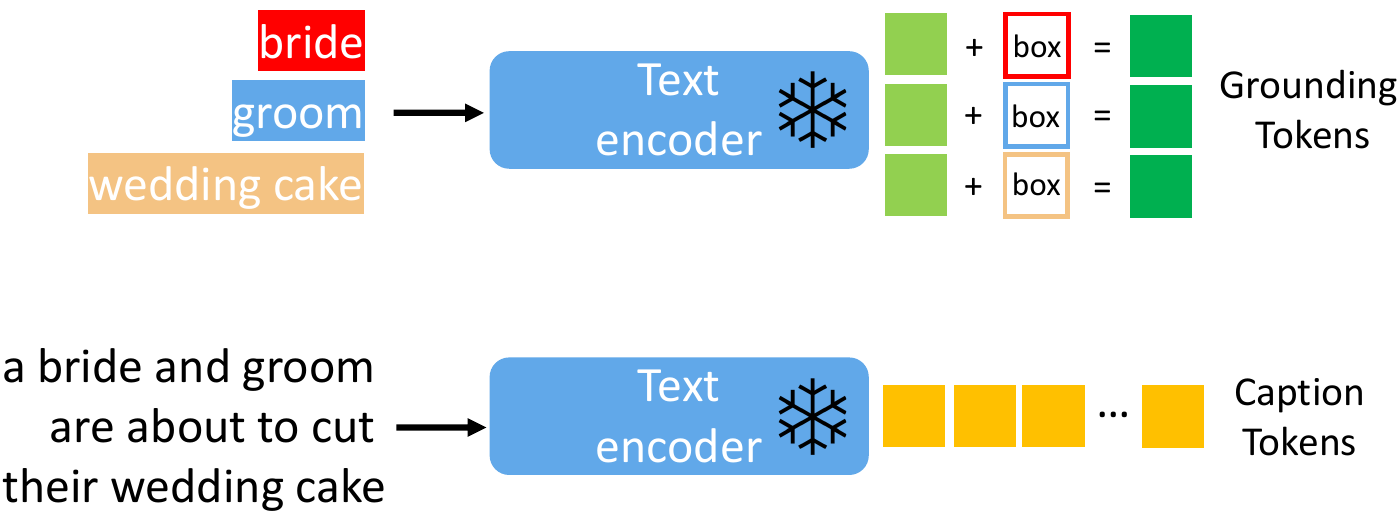}
    \caption{Illustration of grounding token construction process for the bounding box with text case.}
    \label{fig:input_data}
    \vspace{-0.1in}
\end{figure}

\subsection{Grounding Instruction Input} 

For grounded text-to-image generation, there are a variety of ways to ground the generation process via an additional condition. We denote the semantic information of the grounding entity as $e$, which can be described either through text or an example image; and as $\lv$ the grounding spatial configuration described with e.g., a bounding box, a set of keypoints, or an edge map, etc. Note that in certain cases, both semantic and spatial information can be represented with $\lv$ alone (e.g., edge map), in which a single map can represent what objects may be present in the image and where. We define the instruction to a grounded text-to-image model as a composition of the caption and grounded entities:  
\begin{align}\label{eq:data_input}
 ~~~\text{Instruction: }~&	\yv = (\cv, \ev),  ~~~\text{with }~	 \\
  \label{eq:data_input_caption}
	\hspace{-2mm}
 ~~~\text{Caption: }~	
& 
 \cv = [c_1, \cdots,c_L]  \\
 \label{eq:data_input_grounding}
 ~~~\text{Grounding: }~ 
 & 
\ev = [(e_1, \lv_1), \cdots, (e_N, \lv_N)]
\end{align}
where $L$ is the caption length, and $N$ is the number of entities to ground. In this work, we primarily study using bounding box as the grounding spatial configuration $\lv$, because of its large availability and easy annotation for users. For the grounded entity $e$, we mainly focus on using text as its representation due to simplicity. We process both caption and grounding entities as input tokens to the diffusion model, as described in detail below.

\noindent{\textbf{Caption Tokens.}}
The caption $\cv$ is processed in the same way as in LDM. Specifically, we obtain the caption feature sequence (yellow tokens in Figure~\ref{fig:input_data}) using 
 $ \hv^c = [h_1^c, \cdots, h_L^c] = f_{\text{text}}(\cv) $, where $h_{\ell}^c$ is the contextualized text feature for the  ${\ell}$-th word in the caption.

\noindent{\textbf{Grounding Tokens.}} For each grounded text entity denoted with a bounding box, we represent the location information as $\lv = [\alpha_{\min}, \beta_{\min}, \alpha_{\max}, \beta_{\max}]$ with its top-left and bottom-right coordinates. For the text entity $e$, we use the same pre-trained text encoder to obtain its text feature $f_{\text{text}}(e)$ (light green token in Figure~\ref{fig:input_data}), and then fuse it with its bounding box information to produce a grounding token (dark green token in  Figure~\ref{fig:input_data} ):
\begin{equation}
\label{eq:bbox_token}
h^e = \text{MLP}(f_{\text{text}}(e), \text{Fourier}(\lv) )
\end{equation}
where \text{Fourier} is the Fourier embedding~\cite{nerf}, and  $\text{MLP}(\cdot,\cdot)$ is a multi-layer perceptron that first concatenates the two inputs across the feature dimension. 
The grounding token sequence is represented as $\hv^e = [h_1^e, \cdots, h_N^e]$

\noindent{\textbf{From Closed-set to Open-set.}} 
Note that existing layout2img works only deal with a closed-set setting (\eg, COCO categories), as they typically learn a vector embedding $\uv$ per entity, to replace $f_{\text{text}}(e)$ in \eqref{eq:bbox_token}. For a closed-set setting with $K$ concepts, a dictionary of with $K$ embeddings are learned, $\Umat = [\uv_1, \cdots, \uv_K]$. While this non-parametric representation works well in the closed-set setting, it has two drawbacks: (1) The conditioning is implemented as a dictionary look-up over $\Umat$ in the evaluation stage, and thus the model can only ground the observed entities in the generated images, lacking the ability to generalize to ground new entities; (2) No word/phrase is ever utilized in the model condition, and the semantic structure~\cite{jackendoff1992semantic} of the underlying language instruction is missing. In contrast, in our open-set design, since the noun entities are processed by the same text encoder that is used to encode the caption, we find that even when the localization information is limited to the concepts in the grounding training datasets, our model can still generalize to other concepts as we will show in our experiments.

\begin{figure*}[t!]
    \centering
    \includegraphics[width=0.9\textwidth]{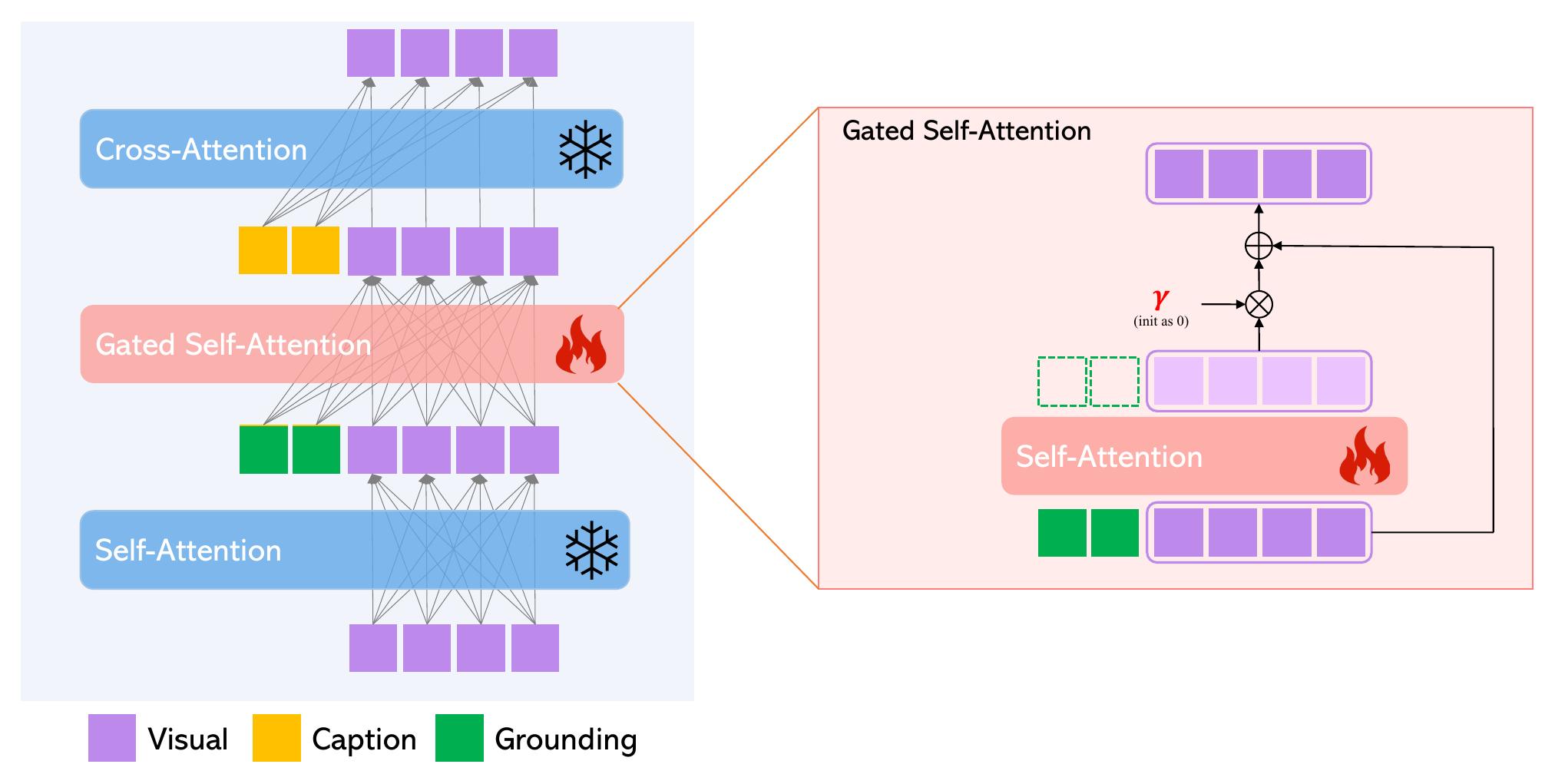}

    \caption{For a pretrained text2img model, the text features are fed into each cross-attention layer. A new gated self-attention layer is inserted to take in the new conditional information.}
    \label{fig:approach}
    \vspace{-0.1in}
\end{figure*}

\noindent{\textbf{Extensions to Other Grounding Conditions.}} 
Note that the proposed grounding instruction in Eq~\eqref{eq:data_input_grounding} is in a general form, though our description thus far has focused on the case of using text as entity $e$ and bounding box as $\lv$ (the major setting of this paper). To demonstrate the flexibility of the \shortname{} framework, we also study additional representative cases which extend the use scenario of Eq~\eqref{eq:data_input_grounding}. 

\begin{itemize}[leftmargin=4.5mm]
\vspace{1mm}
\item {\it Image Prompt.} While language allows users to describe a rich set of entities in an open-vocabulary manner, sometimes more abstract and fine-grained concepts can be better characterized by example images. To this end, one may describe entity $e$ using an image, instead of language.  We use an image encoder to obtain feature $f_{\text{image}}(e)$ which is used in place of $f_{\text{text}}(e)$ in Eq~\eqref{eq:bbox_token} when $e$ is an image.

\vspace{-2mm}
\item {\it Keypoints.} As a simple parameterization method to specify the spatial configuration of an entity, bounding boxes ease the user-machine interaction interface by providing the height and width of the object layout only. One may consider richer spatial configurations such as keypoints for \shortname{}, by parameterizing $\lv$ in Eq~\eqref{eq:data_input_grounding} with a set of keypoint coordinates.  
Similar to encoding boxes, the Fourier embedding~\cite{nerf} can be applied to each keypoint location $\lv=[x,y]$.  

\vspace{-2mm}
\item {\it Spatially-aligned conditions.} To enable more fine-grained controlability, spatially-aligned condition maps can be used, such as edge map, depth map, normal map, and semantic map. In these cases, the semantic information $e$ is already contained within each spatial coordinate $\lv$ of the condition map. A network (\eg conv layers) can be used to encode $\lv$ into $h\times w$ grounding tokens. We also notice that additionally feeding $\lv$ into the first conv layer of the UNet can accelerate training. Specifically, the input to the UNet is $\textsc{concat}( f_l(\lv), \zv_t  )$ where $f_l$ is a simple downsampling network to reduce $\lv$ into the same spatial resolution as $\zv_t$. In this case, the first conv layer of the UNet needs to be trainable.

\vspace{1mm}
\end{itemize}

\noindent Figure~\ref{fig:teaser} shows generated examples for these other grounding conditions. Please refer to the supp for more details.

\subsection{Continual Learning for Grounded Generation}

Our goal is to endow new spatial grounding capabilities to existing large language-to-image generation models. Large diffusion models have been pre-trained on web-scale image-text to gain the required knowledge for synthesizing realistic images based on diverse and complex language instructions. Due to the high pre-training cost and excellent performance, it is important to retain such knowledge in the model weights while expanding the new capability. Hence, we consider to lock the original model weights, and gradually adapt the model by tuning new modules.

\paragraph{Gated Self-Attention.}
We denote $\vv = [v_1, \cdots, v_M]$ as the visual feature tokens of an image. 
The original Transformer block of LDM consists of two attention layers: The self-attention over the visual tokens, followed by cross-attention from caption tokens. By considering the residual connection, the two layers can be written: 
\begin{align}
&	\vv = \vv + \text{SelfAttn}(\vv)	 
\label{eq:ldm_sa}\\
	\hspace{-2mm}
& 
\vv = \vv + \text{CrossAttn}(\vv, \hv^c) 
\label{eq:ldm_ca}
\end{align}

We freeze these two attention layers and add a new gated self-attention layer to enable the spatial grounding ability; see Figure~\ref{fig:approach}.
Specifically, the attention is performed over the concatenation of visual and grounding tokens $[\vv, \hv^e]$: 
\begin{equation}
\label{eq:gated-self-attention}
\vv = \vv + \beta \cdot \tanh(\gamma) \cdot \text{TS}(\text{SelfAttn}([\vv, \hv^e]))
\end{equation}
where $\text{TS}(\cdot) $ is a token selection operation that considers visual tokens only, and $\gamma$ is a learnable scalar which is initialized as 0. $\beta$ is set as 1 during the entire training process and is only varied for scheduled sampling during inference (introduced below) for improved quality and controllability. Note that \eqref{eq:gated-self-attention} is injected in between \eqref{eq:ldm_sa} and \eqref{eq:ldm_ca}. 
Intuitively, the gated self-attention in \eqref{eq:gated-self-attention}  allows visual features to leverage conditional information, and the resulting grounded features are treated as a residual, whose gate is initially set to 0 (due to $\gamma$ being initialized as 0). This also enables more stable training. Note that a similar idea is used in Flamingo~\cite{Alayrac2022FlamingoAV}; however, it uses gated cross-attention, which leads to worse performance in our ablation study.

\paragraph{Learning Procedure.} We adapt the pre-trained model such that grounding information can be injected while all the original components remain intact. By denoting all the new parameters as $\thetav'$, including all gated self-attention layers in Eq~\eqref{eq:gated-self-attention} and MLP in Eq~\eqref{eq:bbox_token}, we use the original denoising objective as in~\eqref{eq:ldm_loss} for model continual learning, based on the grounding instruction input $\yv$: 

\begin{align}\label{eq:grounding_ldm_loss}
\small
\min_{\thetav'} \mathcal{L}_{\text{Grounding}} = \mathbb{E}_{\zv, \epsilonv \sim  \mathcal{N}(\mathbf{0}, \mathbf{I}), t} \big[ \|  \epsilonv -   f_{\{\thetav, \thetav'\}}(\zv_t, t, \yv) \|^2_2 \big].
\end{align}

Why should the model try to use the new grounding information?  Intuitively, predicting the noise that was added to a training image in the reverse diffusion process would be easier if the model could leverage the external knowledge (e.g., each object's location).  Thus, in this way, the model learns to use the additional information while retaining the pre-trained concept knowledge.

\paragraph{Scheduled Sampling in Inference.} The standard inference scheme of \shortname{} is to set $\beta=1$ in~\eqref{eq:gated-self-attention}, and the entire diffusion process is influenced by the grounding tokens. This constant $\beta$ sampling scheme provides overall good performance in terms of both generation and grounding, but sometimes generates lower quality images compared with the original text2img models (e.g., as Stable Diffusion is finetuned on high aesthetic scored images).  To strike a better trade-off between generation and grounding for \shortname{}, we propose a scheduled sampling scheme. As we freeze the original model weights and add new layers to inject new grounding information in training, there is flexibility during inference to schedule the diffusion process to either use both the grounding and language tokens or use only the language tokens of the original model at anytime, by setting different $\beta$ values in~\eqref{eq:gated-self-attention}. Specifically, we consider a two-stage inference procedure, divided by $\tau \in [0,1]$. For a diffusion process with $T$ steps, one can set $\beta$ to 1 at the first $\tau * T$ steps, and set $\beta$ to 0 for the remaining $(1-\tau) * T$ steps:
\begin{align}
\beta = 
\left\{\begin{matrix}
1, & t \le \tau * T ~~~\text{\# Grounded inference stage}~ \\ 
0, & t > \tau * T  ~~~\text{\# Standard inference stage}~~
\end{matrix}\right.
\label{eq:beta_cyclic}
\end{align}

The major benefit of scheduled sampling is improved visual quality as the rough concept location and outline are decided in the early stages, followed by fine-grained details in later stages. It also allows us to extend the model trained in one domain (human keypoint) to other domains (monkey, cartoon characters) as shown in Figure~\ref{fig:teaser}.

\vspace{-1pt}
\section{Experiments}\label{sec:results}
\vspace{-1pt}

We evaluate our model's boxes grounded text2img generation in both the closed-set and open-set settings, and show extensions to other grounding modalities.  We conduct our main quantitative experiments by building upon a pretrained LDM on LAION~\cite{LAION-400M}, unless stated otherwise.

\vspace{-1pt}
\subsection{Closed-set Grounded Text2Img Generation}
\label{sec:closed-set}
\vspace{-1pt}

We first evaluate the generation quality and grounding accuracy of our model in a closed-set setting.  For this, we train and evaluate on the COCO2014~\cite{coco} dataset, which is a standard benchmark used in the text2img literature~\cite{attngan,dfgan,Lafite,DALLE2,Imagen}, and evaluate how the different types of grounding instructions impact our model's performance.

\vspace{-10pt}
\paragraph{Grounding instructions.} We use the following grounding instructions to train our model: 1) COCO2014D: Detection Data.  There are no caption annotations so we use a null caption input~\cite{cfg}. Detection annotations are used as noun-entities. 2) COCO2014CD: Detection + Caption Data. Both caption and detection annotations are used. Note that the noun entities may not always exist in the caption. 3) COCO2014G: Grounding Data.  Given the caption annotations, we use GLIP~\cite{GLIP}, which detects the caption's noun entities in the image, to get pseudo box labels. Please refer to supp for more details about these three types of data.

\begin{table}[t!]
    \begin{center}
        \scriptsize
        \begin{tabular}{ l|c|c|c } 
          \multirow{2}{*}{Model}
            & \multicolumn{2}{c|}{\!Generation: FID  $(\downarrow)$\!}  &   \!\!Grounding: YOLO $(\uparrow)$\!  \\
            & \!Fine-tuned &  \!Zero-shot  & AP/AP$_{50}$/AP$_{75}$ \\
            \hline

            CogView~\cite{ding2021cogview} & -&27.10  & - \\ 
            KNN-Diffusion~\cite{ashual2022knn} & - & 16.66  & - \\    
            DALL-E 2~\cite{DALLE2}  & - & 10.39 & - \\             
            Imagen~\cite{Imagen}  & - & 7.27 & -  \\
            Re-Imagen~\cite{chen2022re}  & 5.25 & 6.88 &   \\            
            Parti~\cite{PARTI}  & 3.20 & 7.23 & -  \\
            LAFITE~\cite{Lafite}  & 8.12  &  26.94 &  -  \\
            LAFITE2~\cite{zhou2022lafite2}  & 4.28  &  8.42 &  -  \\ 

            Make-a-Scene~\cite{MAKEASCENE} & 7.55 &  11.84  &  - \\
            N{\"U}WA~\cite{wu2022nuwa} &  12.90 & - & -\\
            Frido~\cite{frido} & 11.24 &  - &  - \\
            XMC-GAN~\cite{zhang2021crossmodal}&  9.33 & - &  - \\ 
            AttnGAN~\cite{attngan} & 35.49  &  - &  - \\ 
            DF-GAN~\cite{dfgan}  & 21.42  &  - &  -  \\    
            Obj-GAN~\cite{li2019object} & 20.75 & - & - \\
             \hline
            LDM~\cite{LDM}  & - & 12.63 & -  \\            
            LDM* & 5.91 &  11.73 &  0.6 / 2.0 / 0.3 \\
            \rowcolor{Gray}
            \shortname{} (COCO2014CD) & 5.82 & - & 21.7 / 39.0 / 21.7 \\
            \rowcolor{Gray}
            \shortname{} (COCO2014D) & 5.61 & - & \textbf{24.0 / 42.2 / 24.1} \\
            \rowcolor{Gray}
            \shortname{} (COCO2014G) & 6.38 & - & 11.2 / 21.2 /  10.7 \\
            
        \end{tabular}
        \vspace{-0.05in}
	    \caption{Evaluation of image quality and correspondence to layout on COCO2014 val-set. All numbers are taken from corresponding papers, LDM* is our COCO fine-tuned LDM baseline. Here \shortname{} is built upon LDM.}
	    \label{table:text2img_baseline}
	\end{center}
	\vspace{-0.3in}
\end{table}

\vspace{-10pt}
\paragraph{Baselines.} Baseline models are listed in Table~\ref{table:text2img_baseline}. Among them, we also finetune an LDM~\cite{LDM} pretrained on LAION 400M~\cite{LAION-400M} on COCO2014 with its caption annotations, which we denote as LDM*. 

The text2img baselines, as they cannot be conditioned on box inputs, are evaluated on COCO2014C: Caption Data.   

\vspace{-10pt}
\paragraph{Evaluation metrics.} We use the captions and/or box annotations from 30K randomly sampled images to generate 30K images for evaluation. We use {\it FID}~\cite{fid} to evaluate image quality. To evaluate grounding accuracy (\ie correspondence between the input bounding box and generated entity), we use the {\it YOLO score}~\cite{LAMA}.  Specifically, we use a pretrained YOLO-v4~\cite{yolov4} to detect bounding boxes on the generated images and compare them with the ground truth boxes using average precision (AP). Since prior text2img methods do not support taking box annotations as input, it is not fair to compare with them on this metric. Thus, we only report numbers for the fine-tuned LDM as a reference.   

\vspace{-10pt}
\paragraph{Results.} Table~\ref{table:text2img_baseline} shows the results. First, we see that the image synthesis quality of our approach, as measured by FID, is better than most of the state-of-the-art baselines due to rich visual knowledge learned in the pretraining stage. Next, we find that all three grounding instructions lead to comparable FID to that of the LDM* baseline, which is finetuned on COCO2014 with caption annotations. Our model trained using detection annotation instructions (COCO2014D) has the overall best performance. However, when we evaluate this model on COCO2014CD instructions, we find that it has worse performance (FID: 8.2) -- its ability to understand real captions may be limited as it is only trained with the null caption. For the model trained with GLIP grounding instructions (COCO2014G), we actually evaluate it using the COCO2014CD instructions since we need to compute the YOLO score which requires ground-truth detection annotations. Its slightly worse FID may be attributed to its learning from GLIP pseudo-labels. The same reason can explain its low YOLO score (\ie, the model did not see any ground-truth detection annotations during training).

Overall, this experiment shows that: 1) Our model can successfully take in boxes as an additional condition while maintaining image generation quality. 2) All grounding instruction types are useful, which suggests that combining their data together can lead to complementary benefits.

\begin{figure}[t!]
    \centering
    \includegraphics[width=0.45\textwidth]{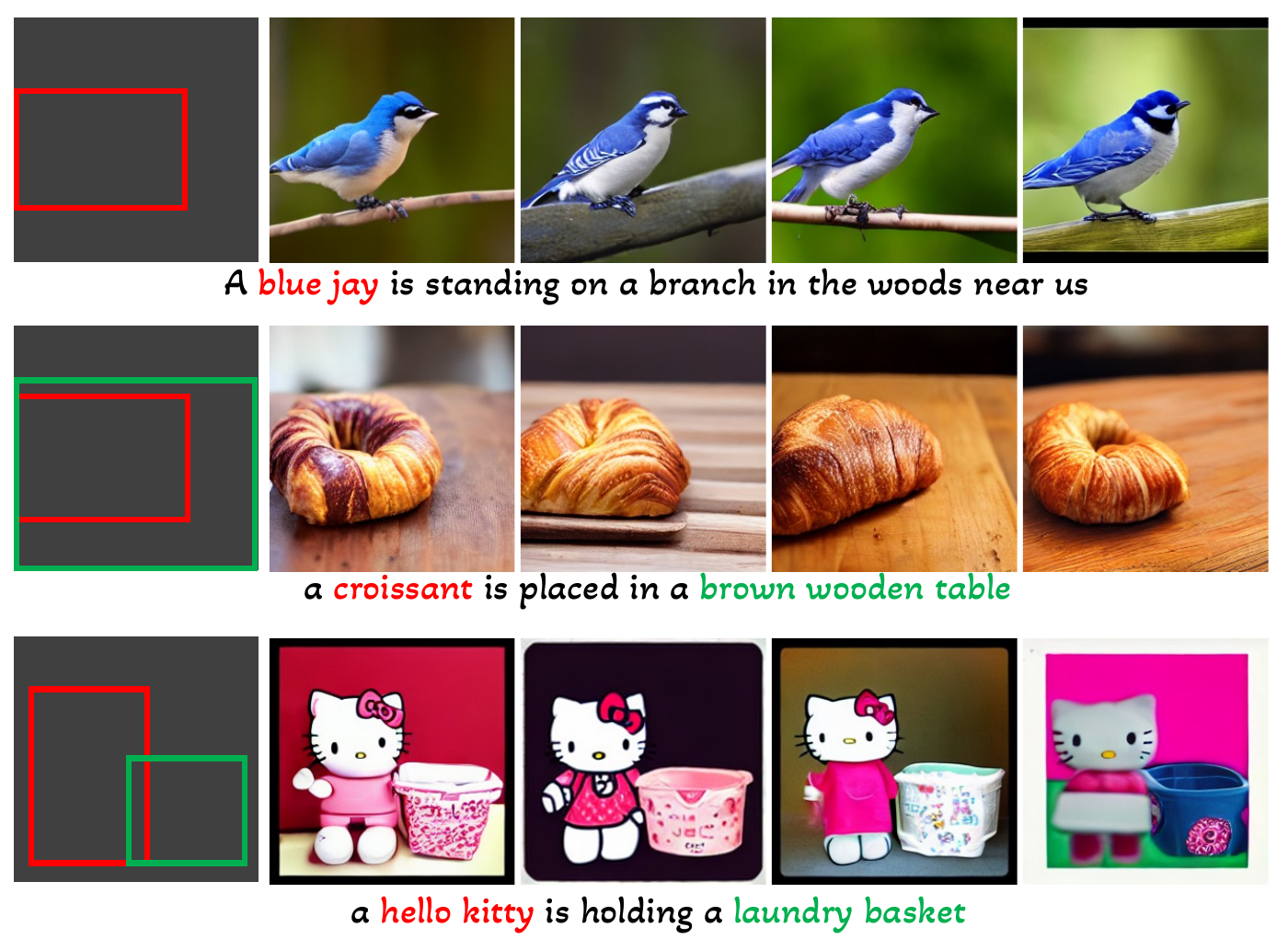}
    \vspace{-5pt}
    \caption{Our model can generalize to  open-world concepts even when only trained using localization annotation from COCO. }
    \label{fig:openwolrd}
    \vspace{-0.21in}
\end{figure}

\vspace{-10pt}
\paragraph{Comparison to Layout2Img generation methods.} Thus far, we have seen that our model correctly learns to use the grounding condition. But how accurate is it compared to methods that are specifically designed for layout2img generation? To answer this, we train our model on COCO2017D, which only has detection annotations. We use the 2017 splits (instead of 2014 as before), as it is the standard benchmark in the layout2img literature. In this experiment, we use the exact same annotation as all layout2img baselines.

\begin{figure}[t!]
    \centering
    \scalebox{1.0}{
    \begin{tabular}{cc}
    \hspace{-5mm}
     \includegraphics[width=.48\linewidth]{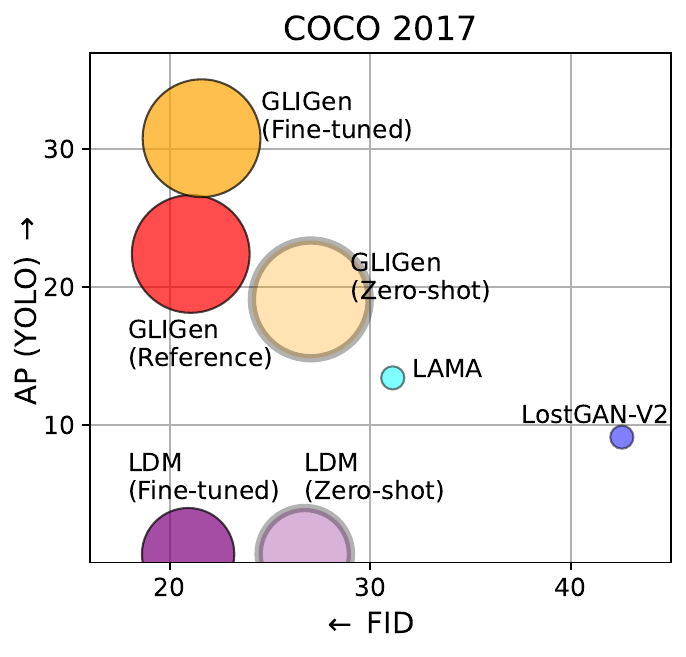}
     & 
     \hspace{-3mm}
     \includegraphics[width=.5\linewidth]{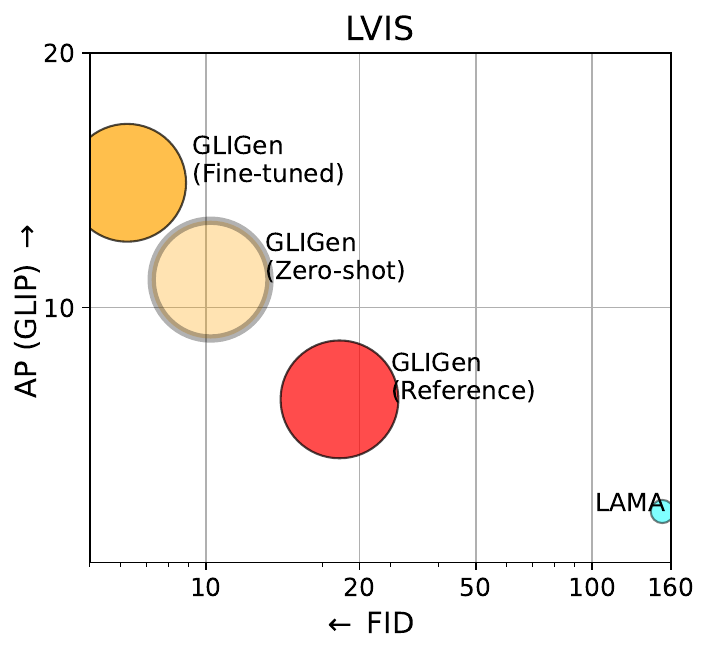}
    \end{tabular}
    }    
    \vspace{-5pt}
    \caption{Performance comparison measured by image generation and grounding quality on COCO2017 (left) and LVIS (right) datasets. \shortname{} is built upon LDM, and continually pre-trained on the joint data of GoldG, O365, SBU, and CC3M. \shortname{} (Reference) is pre-trained on COCO/LVIS only. The circle size indicates the model size. }
    \label{fig:zeroshot}
    \vspace{-0.1in}
\end{figure}

Table~\ref{table:layout2img_baseline} shows that we achieve the state-of-the-art performance for both image quality and grounding accuracy. We believe the core reason is because previous methods train their model from scratch, whereas we build upon a large-scale pretrained generative model with rich visual semantics. Qualitative comparisons are in the supp. We also scale up our training data (discussed later) and pretrain a model on this dataset. Figure~\ref{fig:zeroshot} left shows this model's zero-shot and finetuned results.

\begin{table}[t!]
    \begin{center}
        \scriptsize
        \begin{tabular}{ l|c|c } 
           Model & FID $(\downarrow)$ & YOLO score (AP/AP$_{50}$/AP$_{75}$) $(\uparrow)$  \\
            \hline
            LostGAN-V2~\cite{lostgan2} & 42.55  &   9.1 / 15.3 / 9.8 \\ 
            OCGAN~\cite{ocgan}  & 41.65  &    -  \\
            HCSS~\cite{HCSS}  & 33.68  &    -  \\
            LAMA~\cite{LAMA} & 31.12 &    13.40 / 19.70 / 14.90 \\
            TwFA~\cite{twfa} & 22.15 &    - / 28.20 / 20.12 \\
            \rowcolor{Gray}
            \shortname{}-LDM &       \textbf{21.04} &  \textbf{22.4 / 36.5 / 24.1} \\
 
        \end{tabular}
        \vspace{-0.05in}
	    \caption{Image quality and correspondence to layout are compared with baselines on COCO2017 val-set.} 
	    \label{table:layout2img_baseline}
	\end{center}
	\vspace{-0.1in}
\end{table}

\begin{table}[t!]
    \begin{center}
        \scriptsize
        \begin{tabular}{ 
        l@{\hskip9pt} | c@{\hskip9pt} | c@{\hskip9pt}c@{\hskip7pt}c@{\hskip7pt}c@{\hskip7pt} 
        } 
           Model & Training data  & AP &  AP$_r$ & AP$_c$ & AP$_f$ \\
            \hline
            
            LAMA~\cite{LAMA} & LVIS   & 2.0 & 0.9 & 1.3 & 3.2 \\
            \rowcolor{Gray}
            \shortname{}-LDM   & COCO2014CD   & 6.4 & 5.8 & 5.8 & 7.4  \\
            \rowcolor{Gray}
            \shortname{}-LDM   & COCO2014D   &  4.4 & 2.3 & 3.3 & 6.5  \\
            \rowcolor{Gray}
            \shortname{}-LDM   & COCO2014G   & 6.0 & 4.4 & 6.1 & 6.6  \\ 
            \rowcolor{Gray}
            \shortname{}-LDM   & GoldG,O365   & 10.6 & 5.8 & 9.6 & 13.8  \\
            \rowcolor{Gray}
            \shortname{}-LDM   & GoldG,O365,SBU,CC3M   & 11.1 & 9.0 & 9.8 & 13.4 \\
            \rowcolor{Gray}
            \shortname{}-Stable   & GoldG,O365,SBU,CC3M   & 10.8 & 8.8 & 9.9 & 12.6 \\
            \hline
            Upper-bound   & -   & 25.2 & 19.0 & 22.2 & 31.2  \\

        \end{tabular}
        \vspace{-0.05in}
	    \caption{GLIP-score on LVIS validation set. Upper-bound is provided by running GLIP on real images scaled to 256 $\times$ 256.}
	    \label{table:lvis_ap}
	\end{center}
	\vspace{-0.3in}
\end{table}

\vspace{-1pt}
\subsection{Open-set Grounded Text2Img Generation}
\label{sec:open-set}
\vspace{-1pt}

\begin{figure*}[t!]
    \centering
    \includegraphics[width=0.8\textwidth]{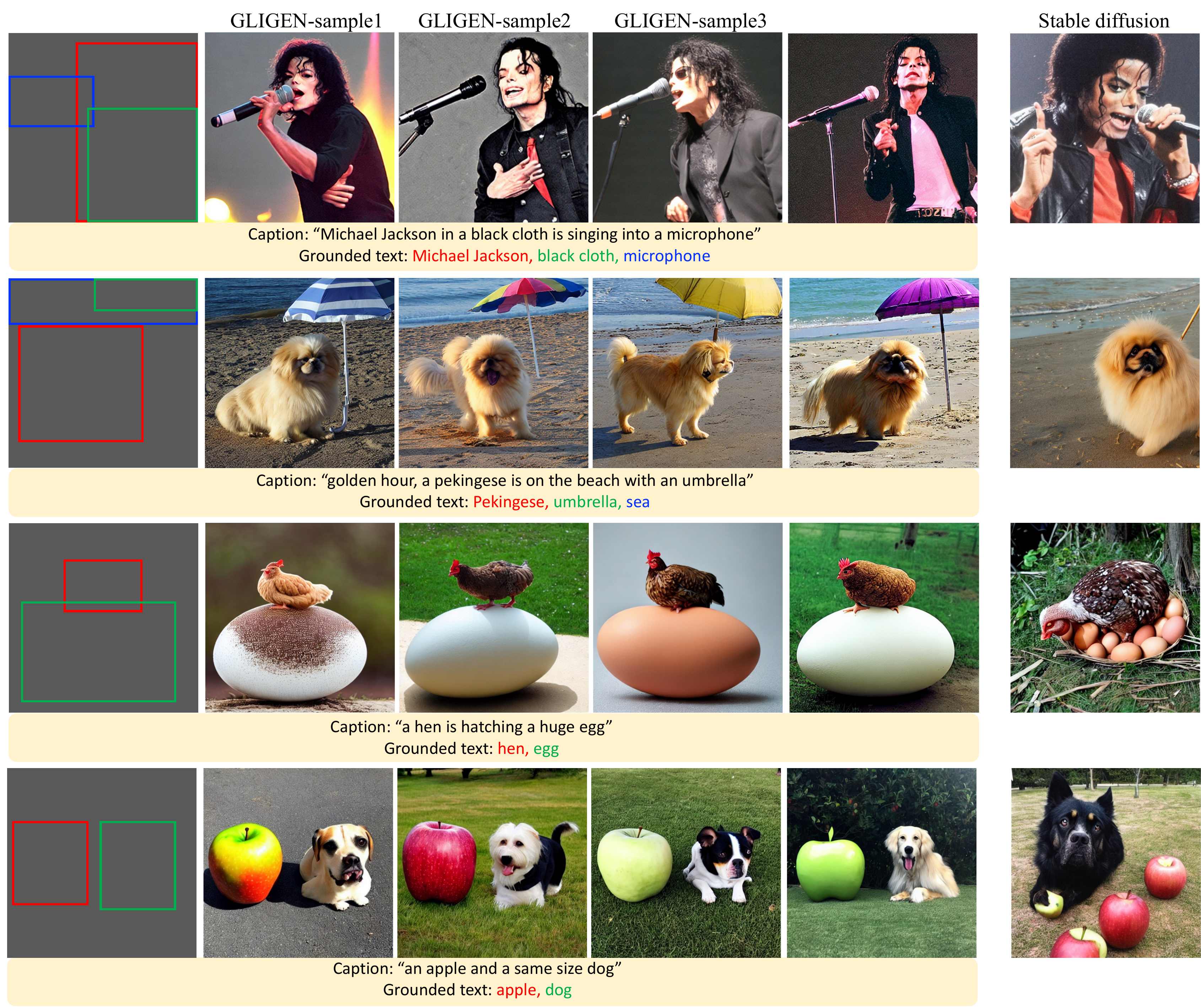}
    \vspace{-10pt}
    \caption{Grounded text2image generation. The baseline lacks grounding ability and can also miss objects \eg ``\texttt{umbrella}'' in a sentence with multiple objects due to CLIP text space, and it also struggles to generate spatially counterfactual concepts.}
    \label{fig:groundtext2img}
    \vspace{-0.1in}
\end{figure*}

\noindent\textbf{COCO-training model.} We first take \shortname{} trained only with the grounding annotations of COCO (COCO2014CD), and evaluate whether it can generate grounded entities beyond the COCO categories.  Figure~\ref{fig:openwolrd} shows qualitative results, where \shortname{} can ground new concepts such as ``\texttt{blue jay}'', ``\texttt{croissant}'' or ground object attributes such as ``\texttt{brown wooden table}'', beyond the training categories. We hypothesize this is because the gated self-attention of \shortname{} learns to re-position the visual features corresponding to the grounding entities in the caption for the ensuing cross-attention layer, and gains generalization ability due to the shared text spaces in these two layers.

We also quantitatively evaluate our model's zero-shot generation performance on LVIS~\cite{lvis}, which contains 1203 long-tail object categories. We use GLIP to predict bounding boxes from the generated images and calculate AP, thus we name it as {\it GLIP score}. We compare to a state-of-the-art model designed for the layout2img task: LAMA~\cite{LAMA}. We train LAMA using the official code on the LVIS training set (in a fully-supervised setting), whereas we directly evaluate our model in a {\it zero-shot task transfer} manner, by running inference on the LVIS val set without seeing any LVIS labels.  Table~\ref{table:lvis_ap} (first 4 rows) shows the results.  Surprisingly, even though our model is only trained on COCO annotations, it outperforms the supervised baseline by a large margin. This is because the baseline, which is trained from scratch, struggles to learn from limited annotations (many of the rare classes in LVIS have fewer than five training samples). In contrast, our model can take advantage of the pretrained model's vast concept knowledge.

\vspace{-10pt}
\paragraph{Scaling up the training data.} We next study our model's open-set capability with much larger training data.  Specifically, we follow GLIP~\cite{GLIP} and train on Object365~\cite{o365} and GoldG~\cite{GLIP}, which combines two grounding datasets: Flickr~\cite{flickr} and VG~\cite{vg}. We also use CC3M~\cite{cc3m} and SBU~\cite{sbu} with grounding pseudo-labels generated by GLIP.  

Table~\ref{table:lvis_ap} shows the data scaling results. As we scale up the training data, our model's zero-shot performance increases, especially for rare concepts. We also try to finetune the model pretrained on our largest dataset on LVIS and demonstrate its performance on Figure~\ref{fig:zeroshot} right. To demonstrate the generality of our method, we also train our model based on the Stable Diffusion model checkpoint using the largest data. We show some qualitative examples in Figure~\ref{fig:groundtext2img} using this model. Our model gains the grounding ability compared to vanilla Stable Diffusion. We notice that Stable Diffusion model may overlook certain objects (``\texttt{umbrella}'' in the second example) due to its use of the CLIP text encoder which tends to focus on global scene properties, and may ignore object-level details~\cite{ediffi}. It also struggles to generate spatially counterfactual concepts. By explicitly injecting entity information through grounding tokens, our model can improve the grounding ability in two ways: the referred objects are more likely to appear in the generated images, and the objects reside in the specified spatial location.

\subsection{Beyond Text Modality Grounding}

\noindent{\textbf{Image grounded generation.}}
One can also use a reference image to represent a grounded entity as discussed previously.  Fig.~\ref{fig:teaser} (b) shows qualitative results, which demonstrate that the visual feature can complement details that are hard to describe by language.

\vspace{-10pt}
\paragraph{Text and image grounded generation.}
Besides using either text or image to represent a grounded entity, one can also keep both representations in one model for more creative generation. Fig.~\ref{fig:teaser} (c) shows text grounded generation with style / tone transfer. For the style reference image, we find that grounding it to an image corner or its edge is sufficient. Since the model needs to generate a harmonious style for the entire image, we hypothesize the self-attention layers may broadcast this information to all pixels, thus leading to consistent style for the entire image.       

\vspace{-10pt}
\paragraph{Keypoints grounded generation.}
We also demonstrate \shortname{} using keypoints for articulate objects control as shown in the Fig.~\ref{fig:teaser} (d). Note that this model is only trained with human keypoint annotations; but it can generalize to other humanoid object due to the scheduled sampling technique we proposed. We also quantitatively study this grounding condition in the supp. 

\vspace{-10pt}
\paragraph{Spatially-aligned condition map grounded generation.} 
Fig.~\ref{fig:teaser} (e-h) demonstrate results for depth map, edge map, normal map, and semantic map grounded generation. These types of conditions allow users to have more fine-grained generation control. See supp for more qualitative results.

\vspace{-2pt}
\subsection{Scheduled Sampling}
\vspace{-2pt}

As stated in Eq.~\eqref{eq:gated-self-attention} and Eq.~\eqref{eq:beta_cyclic}, we can schedule inference time sampling by setting $\beta$ to 1 (use extra grounding information) or 0 (reduce to the original pretrained diffusion model). This can make our model exploit different knowledge at different stages. 

\begin{figure}[t!]
    \centering
    \includegraphics[width=0.48\textwidth]{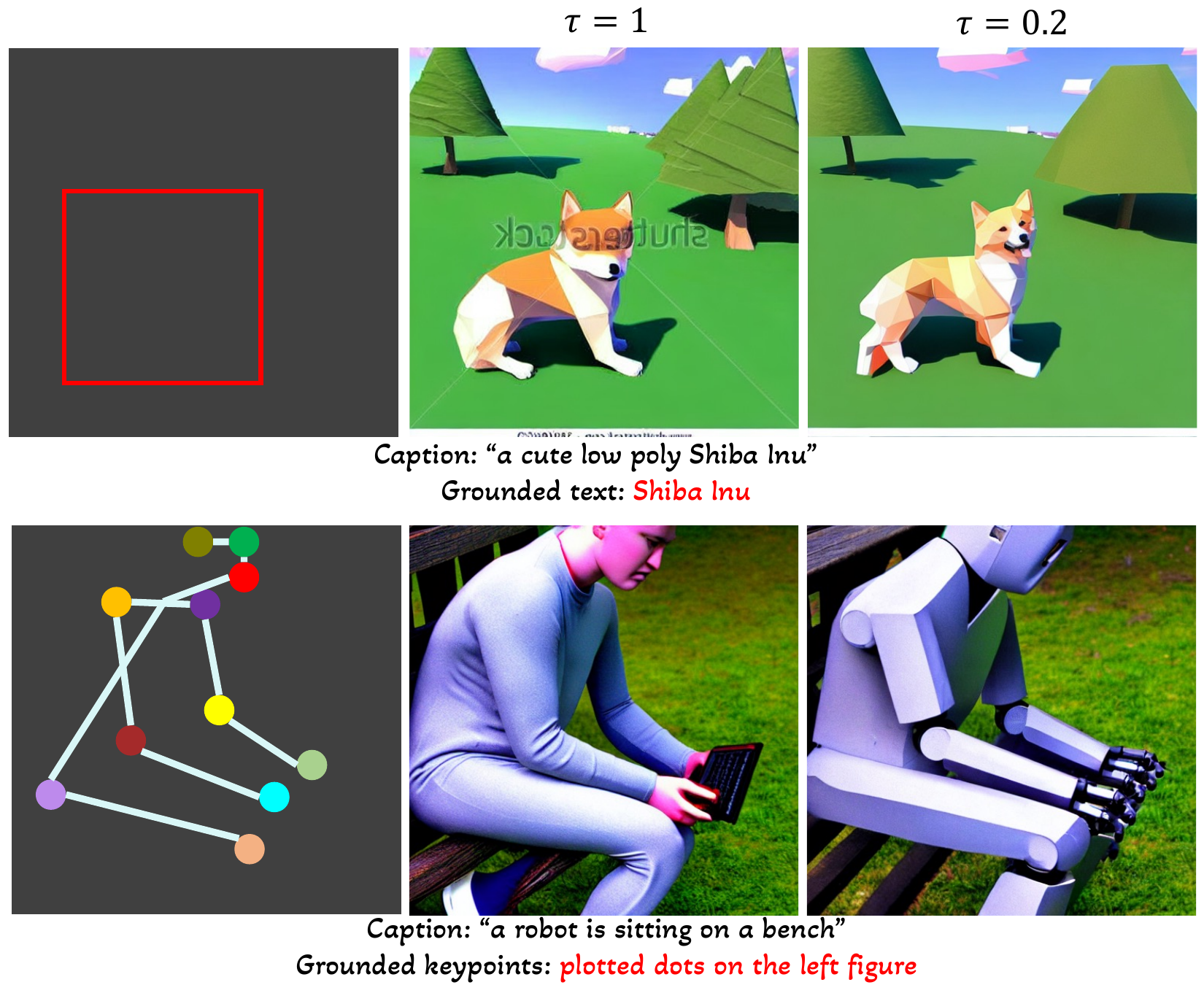}
    \caption{\textbf{Scheduled Samping.} It can improve visual or extend a model trained in one domain (e.g., human) to the others.}
    \label{fig:beta}
    \vspace{-0.1in}
\end{figure}

Fig.~\ref{fig:beta} qualitatively shows the benefits of our scheduled sampling by setting $\tau$ to be 0.2. The images in the same row share the same noise and conditional input. The first row shows that scheduled sampling can be used to improve image quality, as the original Stable Diffusion model is trained with high quality images. The second row shows a generation example by our model trained with COCO human keypoint annotations. Since this model is purely trained with human keypoints, the final result is biased towards generating a human even if a different object (i.e., robot) is specified in the caption. However, by using scheduled sampling, we can extend this model to generate other objects with a human-like shape.

\vspace{-2pt}
\section{Conclusion}\label{sec:conclusion}
\vspace{-2pt}

We proposed \shortname{} for expanding pretrained text2img diffusion models with grounding ability, and demonstrated open-world generalization using bounding boxes as the grounding condition. Our method is simple and effective, and can be easily extended to other conditions such as keypoints, reference images, spatially-aligned conditions (e.g., edge map, depth map, etc). The versatility of \shortname{} makes it a promising direction for advancing the field of text-to-image synthesis and expanding the capabilities of pretrained models in various applications.



\vspace{-3mm}
\paragraph{Acknowledgement.}
This work was supported in part by NSF CAREER IIS2150012, NASA 80NSSC21K0295, and Institute of Information \& communications Technology Planning \& Evaluation(IITP) grants funded by the Korea government(MSIT) (No. 2022- 0-00871, Development of AI Autonomy and Knowledge Enhancement for AI Agent Collaboration) and (No. RS-2022-00187238, Development of Large Korean Language Model Technology for Efficient Pre-training), and Adobe Data Science Research Award.

{\small
\bibliographystyle{ieee_fullname}
\bibliography{egbib}
}

\clearpage
\appendix
\section*{Appendix}

In this supplemental material, we provide more implementation and training details, and then present more results and discussions.

\section{Implementation and training details}

We use the Stable Diffusion model~\cite{LDM} as the example to illustrate our implementation details. 
\vspace{-4mm}
\paragraph{Box Grounding Tokens with Text.} Each grounded text is first fed into the text encoder to get the text embedding (\eg, 768 dimension of the CLIP text embedding in Stable Diffusion). Since the Stable Diffusion uses features of 77 text tokens outputted from the transformer backbone, thus we choose ``\texttt{EOS}'' token feature at this layer as our grounded text embedding. This is because in the CLIP training, this ``\texttt{EOS}'' token feature is chosen and applied a linear transform (one FC layer) to compare with visual feature, thus this token feature should contain whole information about the input text description. We also tried to directly use CLIP text embedding ( after linear projection), however, we notice slow convergence empirically probably due to unaligned space between the grounded text embedding and the caption embeddings.  Following NeRF~\cite{nerf}, we encode bounding box coordinates with the Fourier embedding with output dimension 64. As stated in the Eq~\ref{eq:bbox_token} in the main paper, we first concatenate these two features and feed them into a multi-layer perceptron. The MLP consists of three hidden layers with hidden dimension 512, the output grounding token dimension is set to be the same as the text embedding dimension (\eg, 768 in the Stable Diffusion case). We set the maximum number of grounding tokens to be 30 in the bounding box case.

\vspace{-0.1in}
\paragraph{Box Grounding Tokens with Image.} We use the similar way to get the grounding token for an image. We use the CLIP image encoder (ViT-L-14 is used for the Stable Diffusion) to get an image embedding. We denote the CLIP training objective as maximizing $  (\Pmat_t \hv_t)^{\top}  (\Pmat_i \hv_i) $ (we omit normalization), where $\hv_t$ is ``\texttt{EOS}'' token embedding from the text encoder, $\hv_i$ is ``\texttt{CLS}'' token embedding from the image encoder, and $\Pmat_t$ and $\Pmat_i$ are linear transformation for text and image embedding, respectively. Since $\hv_t$ is the text feature space used for grounded text features, to ease our training, we choose to project image features into the text feature space via $\Pmat_t^{\top} \Pmat_i \hv_i $, and normalized it to 28.7, which is average norm of $\hv_t$ we empirically found. We also set
the maximum number of grounding tokens to be 30. Thus, 60 tokens in total if one keep both image and text as representations for a grounded entity.

\vspace{-0.1in}
\paragraph{Keypoint Grounding Tokens.} The grounding token for keypoint annotations is processed in the same way, except that we also learn $N$ person token embedding vectors $\{\pv_1,\dots,\pv_N\}$ to semantically link keypoints belonging to the same person. This is to deal with the situation in which there are multiple people in the same image that we want to generate, so that the model knows which keypoint corresponds to which person. Each keypoint semantic embedding $\kv_e$  is a learnable vector; the dimension of each person token is set the same as keypoint embedding dimension. The grounding token is calculated by:  
\begin{equation}
\label{eq:keypoint_token}
\hv^e = \text{MLP}(\kv_e +\pv_j, \text{Fourier}(\lv) )
\end{equation}
where $\lv$ is the $x,y$ location of each keypoint and $\pv_j$ is the person token for the $j$'th person. In practice, we set $N$ as 10, which is the maximum number of persons allowed to be generated in each image. Thus, we have 170 tokens in the COCO dataset (\ie, 10*17; 17 keypoint annotations for each person).

\begin{figure}
    \centering
    \includegraphics[width=0.40\textwidth]{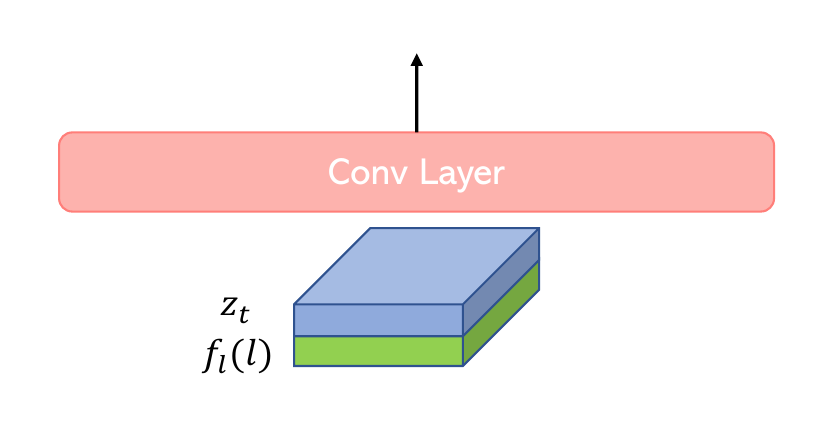}
    \caption{Additional grounding input is fed into the Unet input for spatially aligned conditions.}
    \label{fig:unet_input}
\end{figure}

\paragraph{Tokens for Spatially Aligned Condition.}
This type of condition includes edge map, depth map, semantic map, and normal map, etc; they can be represented as $C\times H \times W$ tensor. We resize spatial size into $256 \times 256$ and use the convnext-tiny~\cite{liu2022convnet} as the backbone to output a feature with spatial size as $8 \times 8$, which then is flattened into 64 grounding tokens. We notice that it can help training faster if we also provide the grounding condition $l$ into the Unet input. As shown in the Figure~\ref{fig:unet_input}, in this case, the input is $\textsc{concat}( f_l(\lv), \zv_t  )$ where $f_l$ is a simple downsampling network to reduce $\lv$ into the same spatial dimension as $\zv_t$, which is the noisy latent code at the time step $t$ ($64\times64$ for the Stable Diffusion). In this case, the first conv layer of Unet needs to be trainable.

\vspace{-4mm}
\paragraph{Gated Self-Attention Layers.}
Our inserted self-attention layer is the same as the original diffusion model self-attention layer at each Transformer block, except that we add one linear projection layer which converts the grounding token into the same dimension as the visual token. For example, in the first layer of the down branch of the UNet~\cite{unet}, the projection layer converts grounding token of dimension 768 into 320 (which is the image feature dimension at this layer), and visual tokens are concatenated with the grounding tokens as the input to the gated attention layer.

\begin{figure}
    \centering
    \includegraphics[width=0.45\textwidth]{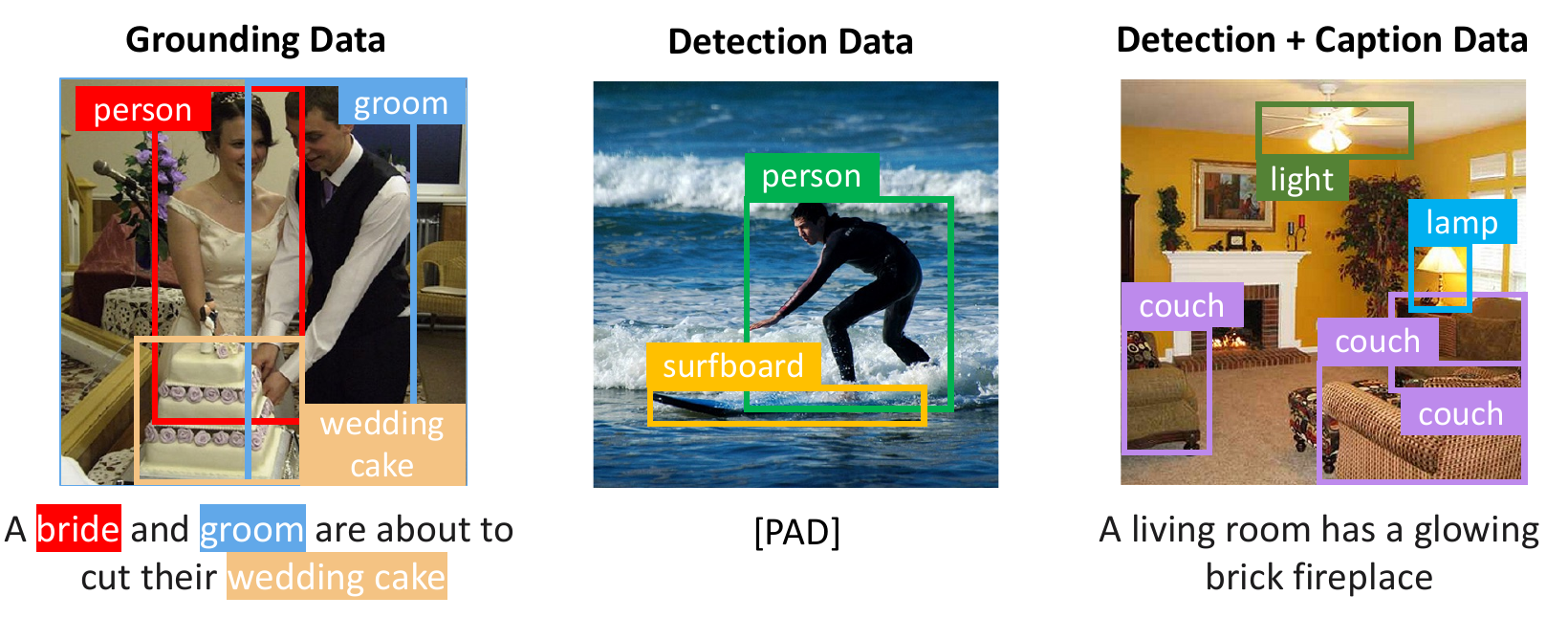}
    \caption{Three different types of grounding data for box.}
    \label{fig:data_type}
\end{figure}

\vspace{-4mm}
\paragraph{Training Details.} For all COCO related experiments (Sec~\ref{sec:closed-set} in the main paper), we train LDM with batch size 64 using 16 V100 GPUs for 100k iterations. In the scaling up training data experiment (in Sec~\ref{sec:open-set} of the main paper), we train for 400k iterations for LDM, but 500K iterations with batch size of 32 for the Stable diffusion modeL For all training, we use learning rate of 5e-5 with Adam~\cite{adam}, and use warm-up for the first 10k iterations. We randomly drop caption and grounding tokens with 10\% probability for classifier-free guidance~\cite{cfg}.

\paragraph{Data Details.} 

In the main paper Sec~\ref{sec:closed-set}, we study three different types of data for box grounding. The training data requires both text $\cv$ and grounding entity $\ev$ as the full condition. In practice, we can relax the data requirement by considering a more flexible input, \ie the three types of data shown in Figure~\ref{fig:data_type}.  $(i)$ {\it Grounding data}. Each image is associated with a caption describing the whole image; noun entities are extracted from the caption, and are labeled with bounding boxes. Since the noun entities are taken directly from the natural language caption, they can cover a much richer vocabulary which will be beneficial for open-world vocabulary grounded generation. $(ii)$ {\it Detection data}. Noun-entities are pre-defined closed-set categories (\eg, 80 object classes in COCO~\cite{coco}). In this case, we choose to use a null caption token as introduced in classifier-free guidance~\cite{cfg} for the caption. The detection data is of larger quantity (millions) than the grounding data (thousands), and can therefore greatly increase overall training data. $(iii)$ {\it Detection and caption data}. Noun entities are same as those in the detection data, and the image is described separately with a text caption. In this case, the noun entities may not exactly match those in the caption. For example, in Figure~\ref{fig:data_type}, the caption only gives a high-level description of the living room without mentioning the objects in the scene, whereas the detection annotation provides more fine-grained object-level details.

\section{Ablation Study}

\paragraph{Ablation on gated self-attention.} As shown in the main paper Figure~\ref{fig:approach} and Eq~\ref{eq:gated-self-attention}, our approach uses gated self-attention to absorb the grounding instruction. We can also consider gated cross-attention~\cite{Alayrac2022FlamingoAV}, where the query is the visual feature, and the keys and values are produced using the grounding condition. We ablate this design on COCO2014CD data using LDM. Compare with the Table~\ref{table:text2img_baseline} the main paper, we can find that it leads to similar FID: 5.8, but worse YOLO AP: 16.6 (compared to 21.7 for self-attention in the Table). This shows the necessity of information sharing among the visual tokens, which exists in self-attention but not in cross-attention. 

\paragraph{Ablation on null caption.} We choose to use the null caption when we only have detection annotations (COCO2014D). An alternative scheme is to simply combine all noun entities into a sentence;~\eg, if there are two cats and a dog in an image, then the pseudo caption can be: ``\texttt{cat, cat, dog}''. In this case, the FID becomes worse and increases to 7.40 from 5.61 (null caption, refer to main paper table 1). This is likely due to the pretrained text encoder never having encountered this type of unnatural caption during LDM training. A solution would be to finetune the text encoder or design a better prompt, but this is not the focus of our work.

\paragraph{Ablation on fourier embedding.} In Eq~\ref{eq:bbox_token}, we replace the Fourier embedding with MLP embedding and conduct an experiment using COCO2014CD data format (Table~\ref{table:text2img_baseline}) . In this case, the image quality 
(FID) is similar: Fourier/MLP: 5.82/\textbf{5.80}; however, the layout correspondence (YOLO AP) is much worse: Fourier/MLP: \textbf{21.7}/3.2.

\section{Grounded inpainting}

\begin{figure}[t!]
    \centering
    \includegraphics[width=0.48\textwidth]{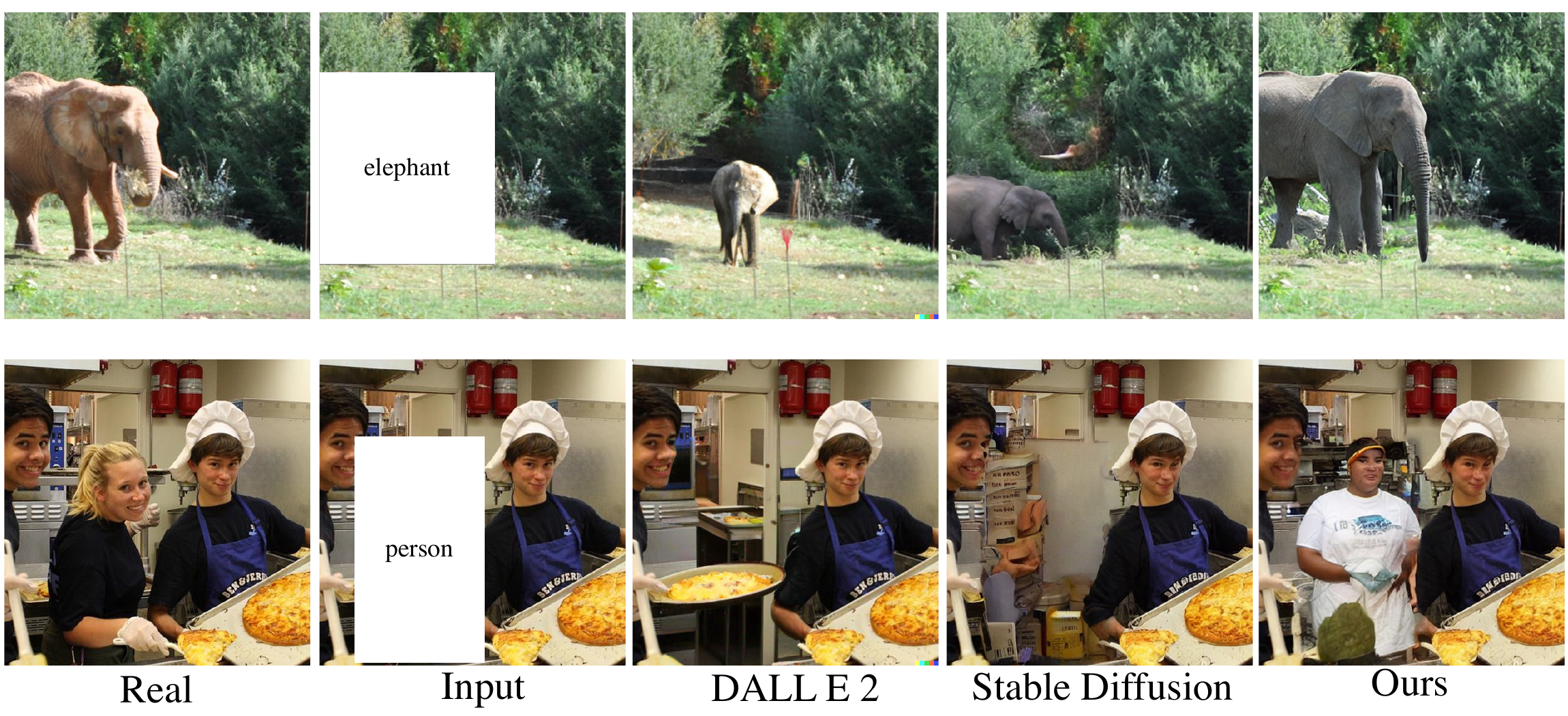}
    \vspace{-0.18in}
    \caption{\textbf{Inpainting results.} Existing text2img diffusion models may generate objects that do not tightly fit the masked box or miss an object if the same object already exists in the image.}
    \label{fig:inpaint}
    \vspace{0.02in}
\end{figure}
\begin{table}[t!]
    \begin{center}
        \begin{tabular}{ l| c|c|c } 
           & 1\%-3\% & 5\%-10\% & 30\%-50\% \\
            \hline
            LDM~\cite{LDM} & 25.9  & 23.4 & 14.6 \\ 
            \rowcolor{Gray}
            \shortname{}-LDM        & \textbf{29.7}  & \textbf{30.9} & \textbf{25.6}   \\
             \hline
            Upper-bound & 41.7  & 43.4 & 45.0 \\   
        \end{tabular}
        \vspace{-0.05in}
	    \caption{Inpainting results (YOLO AP) for different size of objects.}
	    \label{table:inpainting}
	\end{center}
	\vspace{-0.3in}
\end{table}

\begin{figure}
    \centering
    \includegraphics[width=0.42\textwidth]{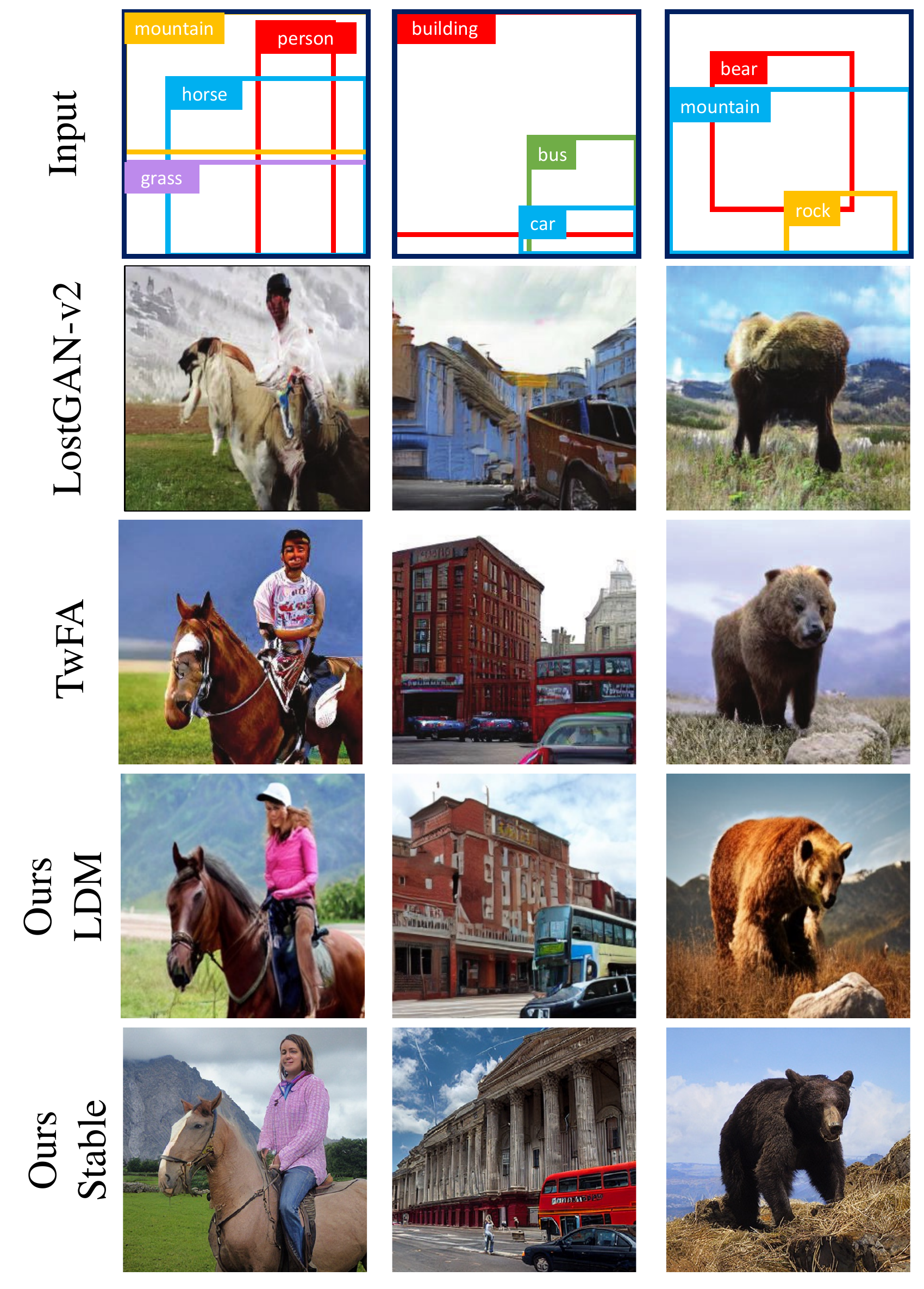}
    \caption{Layout2img comparison. Our model generates better quality images, especially when using stable diffusion. Baseline images are all copied from TwFA~\cite{twfa}}
    \label{fig:supp_layout2img}
    \vspace{-0.1in}
\end{figure}

\subsection{Text Grounded Inpainting} 

Like other diffusion models, \shortname{} can also work for the inpainting task by replacing the known region with a sample
from $q(z_t|z_0)$ after each sampling step, where $z_0$ is the latent representation of an image~\cite{LDM}. One can ground text descriptions to missing regions, as shown in Figure~\ref{fig:inpaint}. In this setting, however, one may wonder, can we simply use a vanilla text-to-image diffusion model such Stable Diffusion or DALLE2 to fill the missing region by providing the object name as the caption? What are the benefits of having extra grounding inputs in such cases? To answer this, we conduct the following experiment on the COCO dataset: for each image, we randomly mask one object. We then let the model inpaint the missing region. We choose the missing object with three different size ratios with respect to the image: small (1\%-3\%), median (5\%-10\%), and large (30\%-50\%). 5000 images are used for each case.

Table~\ref{table:inpainting} demonstrates that our inpainted objects more tightly occupy the missing region (box) compared to the baselines. Fig.~\ref{fig:inpaint}  provides examples to visually compare the inpainting results (we use Stable Diffusion for better quality). The first row shows that baselines' generated objects do not follow the provided box. The second row shows that when the missing category is already present in the image, they may ignore the caption. This is understandable as baselines are trained to generate a \emph{whole} image following the caption. Our method may be more favorable for editing applications, where a user might want to generate an object that fully fits the missing region or add an instance of a class that already exists in the image.

\begin{figure}[]
    \centering
    \includegraphics[width=0.48\textwidth]{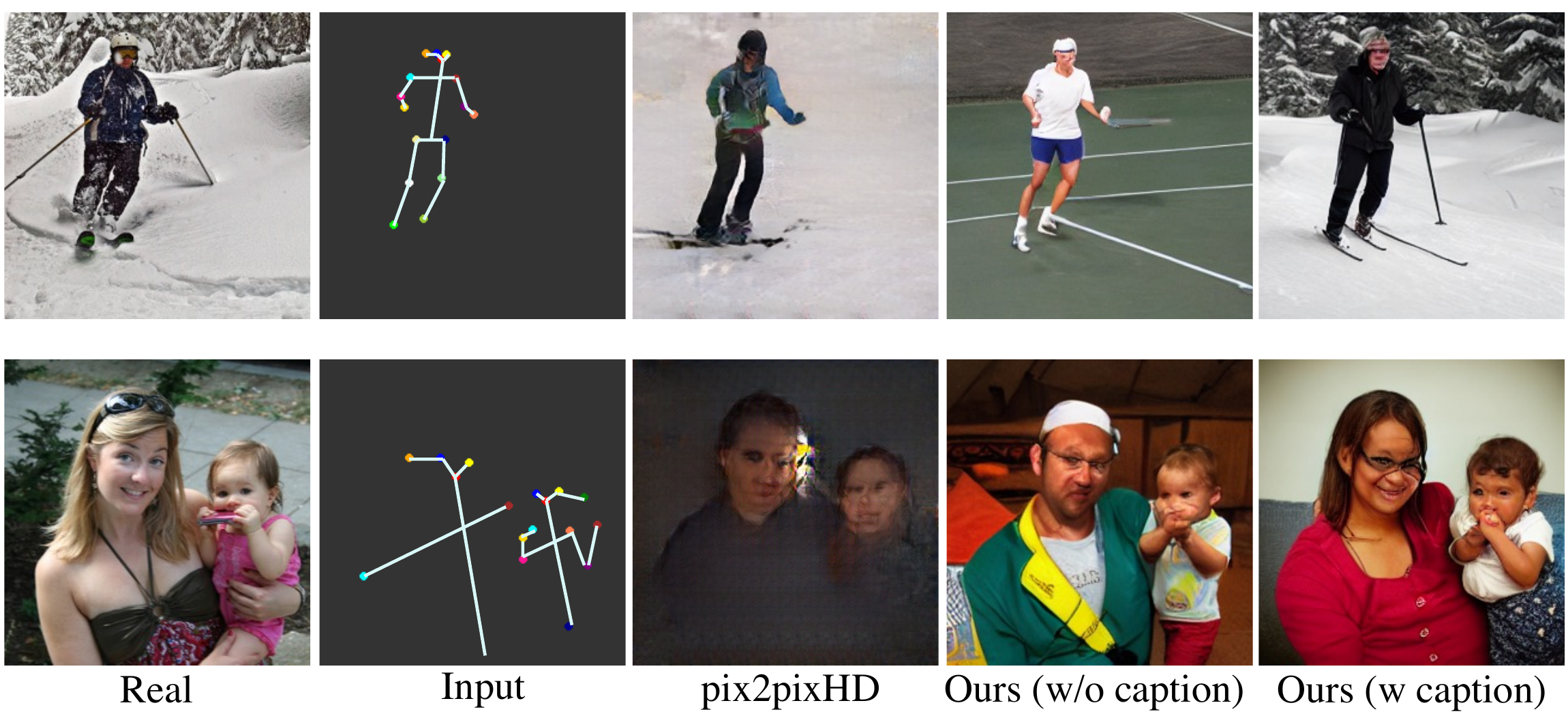}
    \vspace{-12pt}
    \caption{\textbf{Keypoint results.} Our model generates higher quality images conditioned on keypoints, and it allows to use caption to specify details such as scene or gender.}
    \label{fig:keypoint}
    \vspace{-0.05in}
\end{figure}

\begin{table*}
    \begin{center}
    \scriptsize
        \begin{tabular}{ 
        lllc cccc 
        } 
        \toprule
           Model &  Pre-training data & Traing data & FID & AP &  AP$_r$ & AP$_c$ & AP$_f$ \\
            \midrule
            
            LAMA~\cite{LAMA} & --  & LVIS & 151.96 & 2.0 & 0.9 & 1.3 & 3.2 \\
            \rowcolor{Gray}
            \shortname{}-LDM   & COCO2014CD  & -- & 22.17  & 6.4 & 5.8 & 5.8 & 7.4  \\
            \rowcolor{Gray}
            \shortname{}-LDM   & COCO2014D  & --  & 31.31  &  4.4 & 2.3 & 3.3 & 6.5  \\
            \rowcolor{Gray}
            \shortname{}-LDM   & COCO2014G & --   & 13.48 & 6.0 & 4.4 & 6.1 & 6.6  \\ 
            \rowcolor{Gray}
            \shortname{}-LDM   & GoldG,O365  & -- & 8.45 & 10.6 & 5.8 & 9.6 & 13.8  \\
            \rowcolor{Gray}
            \shortname{}-LDM   & GoldG,O365,SBU,CC3M  & -- & 10.28  & 11.1 & 9.0 & 9.8 & 13.4 \\
            \rowcolor{Gray}
            \shortname{}-LDM   & GoldG,O365,SBU,CC3M & LVIS & \textbf{6.25} & \textbf{14.9} & \textbf{10.1} & \textbf{12.8} & \textbf{19.3} \\
 
            \midrule
            Upper-bound   & --   & -- & -- & 25.2 & 19.0 & 22.2 & 31.2  \\
        \bottomrule
        \end{tabular}
	    \caption{GLIP-score on LVIS validation set. Upper-bound is provided by running GLIP on real images scaled to 256 $\times$ 256.}
	    \label{table:supp_lvis_ap}
	\end{center}
\end{table*}

\subsection{Image Grounded Inpainting} 

\begin{figure*}[t!]
    \centering
    \includegraphics[width=0.90\textwidth]{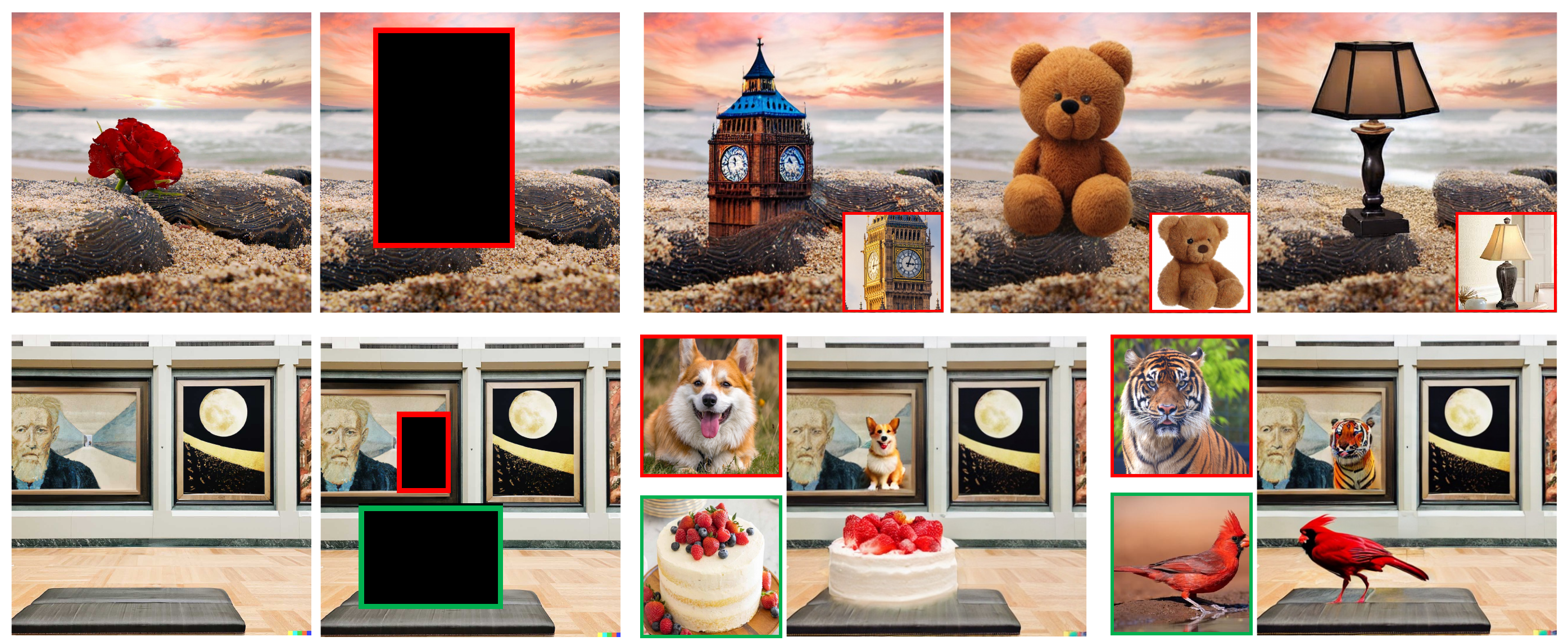}
    \vspace{-10pt}
    \caption{\textbf{Image grounded Inpainting.} One can use reference images to ground holes they want to fill in.}
    \label{fig:image_grounded_inpainting}
    \vspace{-0.1in}
\end{figure*}

As we previously demonstrated, one can ground text to missing region for inpainting, one can also ground reference images to missing regions. Figure~\ref{fig:image_grounded_inpainting} shows inpainting results grounded on reference images. To remove boundary artifacts, we follow GLIDE~\cite{GLIDE}, and modify the first conv layer by adding 5 extra channels (4 for $z_0$ and 1 for inpainting mask) and make them trainable with the new added layers.

\begin{table}[]
    \begin{center}
        \scriptsize
        \begin{tabular}{ 
        l@{\hskip9pt} | c@{\hskip9pt} | c@{\hskip9pt}c@{\hskip7pt}c@{\hskip7pt}c@{\hskip7pt} 
        } 
           Model & FID  & AP &  AP$_{50}$ & AP$_{75}$ \\
            \hline
            
            pix2pixHD~\cite{p2phd}   & 142.4   & 15.8 & 33.7 & 13.0  \\
            \rowcolor{Gray}
            \shortname{} (w/o caption)  & 31.02   & \textbf{31.8} & \textbf{53.5} & \textbf{31.0}   \\ 
            \rowcolor{Gray}
            \shortname{} (w caption)    & \textbf{27.34}   & 31.5 & 52.9 & \textbf{31.0}   \\
            \hline
            Upper-bound   & -   & 62.4 & 75.0 & 65.9  \\

        \end{tabular}
        \vspace{-0.05in}
	    \caption{Conditioning with Human Keypoints evaluated on COCO2017 validation set. Upper-bound is calculated on real images scaled to 256 $\times$ 256.}
	    \label{table:keypoints}
	\end{center}
	\vspace{-0.3in}
\end{table}

\section{Study for Keypoints Grounding}
\vspace{-5pt}

Although we have thus far demonstrated results with bounding boxes, our approach has flexibility in the grounding condition that it can use for generation. To demonstrate this, we next evaluate our model with another type of grounding condition: human keypoints. We use the COCO2017 dataset. We compare with pix2pixHD~\cite{p2phd}, a classic image-to-image translation model. Since pix2pixHD does not take captions as input, we train two variants of our model: one uses COCO captions, the other does not. In the latter case, null caption is used as input to the cross-attention layer for a fair comparison. 

Fig.~\ref{fig:keypoint} shows the qualitative comparison. Clearly, our method generates much better image quality. For our model trained with captions, we can also specify other details such as the scene (``\texttt{\small A person is skiing down a snowy hill}'') or person's gender (``\texttt{\small A woman is holding a baby}''). These two inputs complement each other and can enrich a user's controllability for image creation. We measure keypoint correspondence (similar to the YOLO score for boxes) by running a MaskRCNN~\cite{maskrcnn} keypoint detector on the generated images. Both of our model variants produce similar results; see Table~\ref{table:keypoints}.

\section{Additional quantitative results}

In this section, we show more studies with our pretrained model using our largest data (GoldG, O365, CC3M, SBU). We had reported this model's zero-shot performance on LVIS~\cite{lvis} in the main paper Table~\ref{table:lvis_ap}. Here we finetune this model on LVIS, and report its GLIP-score in Table~\ref{table:supp_lvis_ap}. Clearly, after finetuning, we show much more accurate generation results,  surpassing the supervised baseline LAMA~\cite{LAMA} by a large margin. 

Similarly, we also test this model's zero-shot performance on the COCO2017 val-set, and its finetuning results are in Table~\ref{table:supp_layout2img_baseline}. The results show the benefits of pretraining which can largely improve layout correspondence performance.

\begin{table}[t]
    \begin{center}
    \footnotesize
        \begin{tabular}{ l|c|ccc } 
            \toprule
           & & \multicolumn{3}{c}{YOLO score} \\
           Model & FID & AP & AP$_{50}$ &AP$_{75}$ \\
            \midrule
            LostGAN-V2~\cite{lostgan2} & 42.55  &   9.1 & 15.3 & 9.8 \\ 
            OCGAN~\cite{ocgan}  & 41.65  &  \multicolumn{3}{c}{--}\\
            HCSS~\cite{HCSS}  & 33.68  & \multicolumn{3}{c}{--}  \\
            LAMA~\cite{LAMA} & 31.12 &  13.40 & 19.70 & 14.90 \\
            TwFA~\cite{twfa} & 22.15 &  -- & 28.20 & 20.12 \\
            \rowcolor{Gray}
            \shortname{}-LDM &       \textbf{21.04} &  22.4 & 36.5 & 24.1 \\
            \midrule
            
            \multicolumn{5}{l}{\textit{After pretrain on GoldG,O365,SBU,CC3M}} \\
            \rowcolor{Gray}
            \shortname{}-LDM ({\it zero-shot}) & 27.03 & 19.1 & 30.5 & 20.8 \\
            \rowcolor{Gray}
            \shortname{}-LDM ({\it finetuned}) & 21.58 & \textbf{30.8} & \textbf{42.3} & \textbf{35.3} \\
            \bottomrule
 
        \end{tabular}
	    \caption{Image quality and correspondence to layout are compared with baselines on COCO2017 val-set.} 
	    \label{table:supp_layout2img_baseline}
	\end{center}
\end{table}

\begin{figure*}[]
    \centering
    \includegraphics[width=0.90\textwidth]{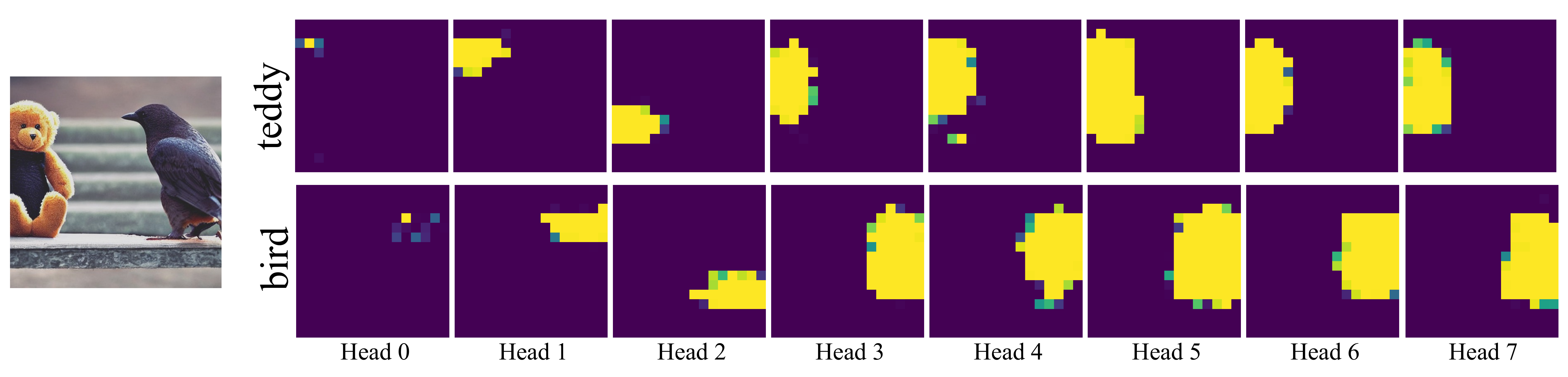}
    \vspace{-10pt}
    \caption{\textbf{Attention maps in one gated self-attention layer.} The visualization results are from the sample at the first time step (i.e., Gaussian noise) in the middle layer of the Unet.}
    \label{fig_supp:attn_map}
    \vspace{-0.1in}
\end{figure*}

\begin{figure}[]
    \centering
    \includegraphics[width=0.32\textwidth]{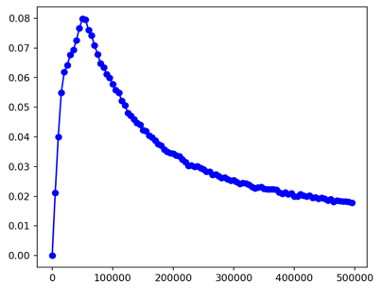}
    \vspace{-10pt}
    \caption{learnable $\gamma$ in the gated self attention layer in the middle of Unet changes during the training progress.}
    \label{fig_supp:gamma_change}
    \vspace{-0.2in}
\end{figure}

\section{Analysis on \shortname{}}

To have a better understanding of \shortname{}, we choose to study the box grounded model. Specifically, we try to visualize attention maps within gated self-attention layer and how does the learnable $\gamma$ in Eq~\ref{eq:gated-self-attention} change during the training process.    

In the Figure~\ref{fig_supp:attn_map}, we first show a generation result using two grounding tokens (teddy bear; bird). Next to it, we visualize the attention maps of our added layers between the visual features and two grounding tokens for all 8 heads for one middle layer in the UNet. Even for the first sampling step (input is Gaussian noise), the visual feature starts to attend to the grounding tokens with correct spatial correspondence. This correspondence fades away in later sampling steps (which is aligned with our `scheduled sampling technique' where we find rough layout is decided in the early sample steps). 

We also find the attention maps for the beginning layers of the UNet to be less interpretable for all sample steps. We hypothesize that this is due to the lack of positional embedding for visual tokens, whereas position information can be leaked into later layers through zero padding via Conv layers. This might suggest that adding positional embedding for diffusion model pretraining (e.g., Stable Diffusion model training) could benefit downstream adaptation. 

The Figure~\ref{fig_supp:gamma_change} shows how the learned $\gamma$ at this layer (Eq~\ref{eq:gated-self-attention}) changes during training. We empirically find the model starts to learn the correspondence around 60-70k iterations (around the peak in the plot). We hypothesize the model tries to focus on learning spatial correspondence at the beginning of training, then tries to finetune and dampen the new layers' contribution so that it can focus on image quality and details as the original weights are fixed.

\section{More qualitative results}

We show qualitative comparisons with layout2img baselines in  Figure~\ref{fig:supp_layout2img}, which complements the results in Sec~\ref{sec:closed-set} of the main paper. The results show that our model has comparable image quality when built upon LDM, but has more visual appeal and details when built upon the Stable Diffusion model.  

\begin{figure*}[t]
    \centering
    \includegraphics[width=0.90\textwidth]{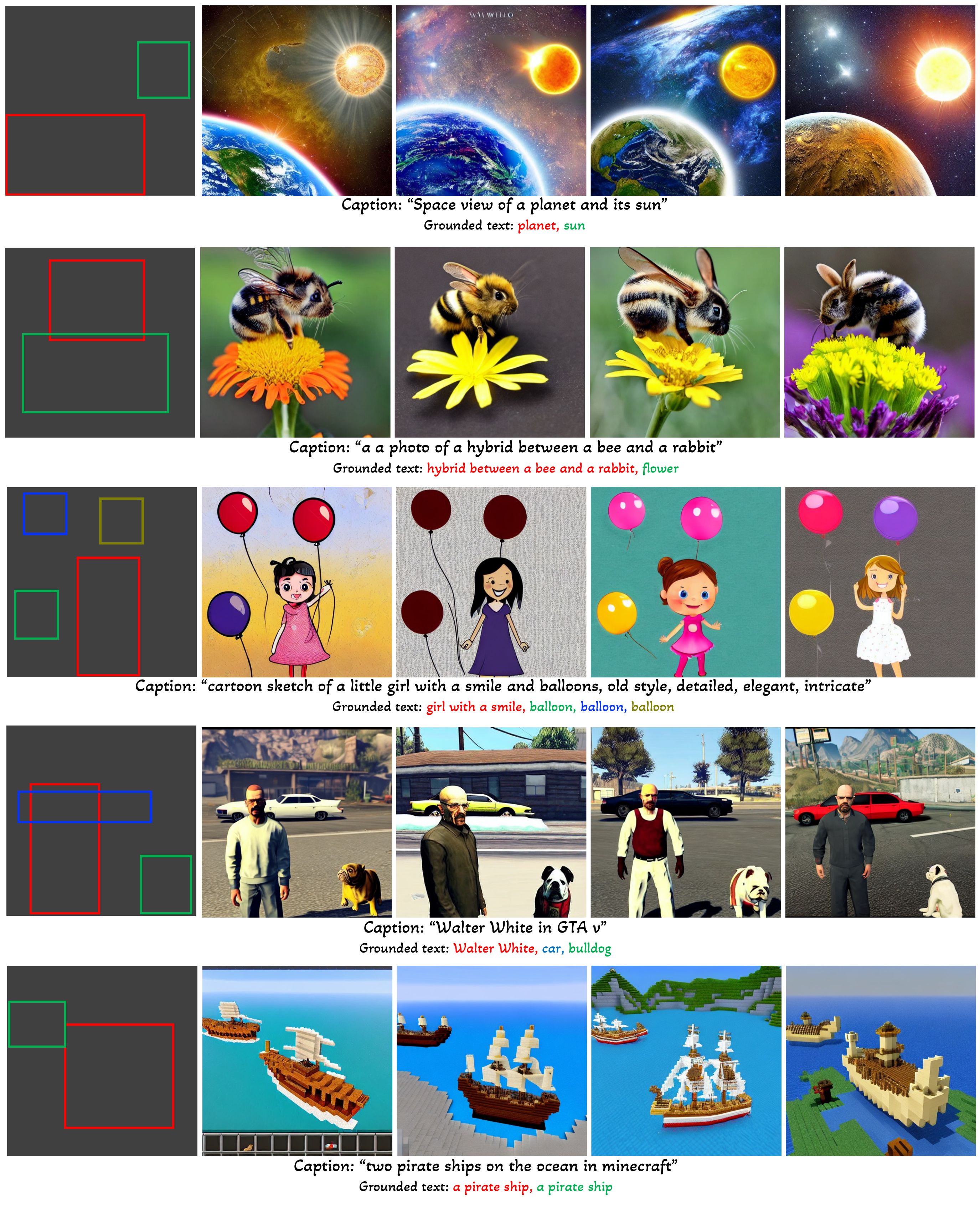}
    \caption{Bounding box grounded text2image generation. Our model can ground noun entities in the caption for controllable image generation}
    \label{fig:supp_groundtext2img}
    \vspace{-0.1in}
\end{figure*}

\begin{figure*}[t]
    \centering
    \includegraphics[width=0.90\textwidth]{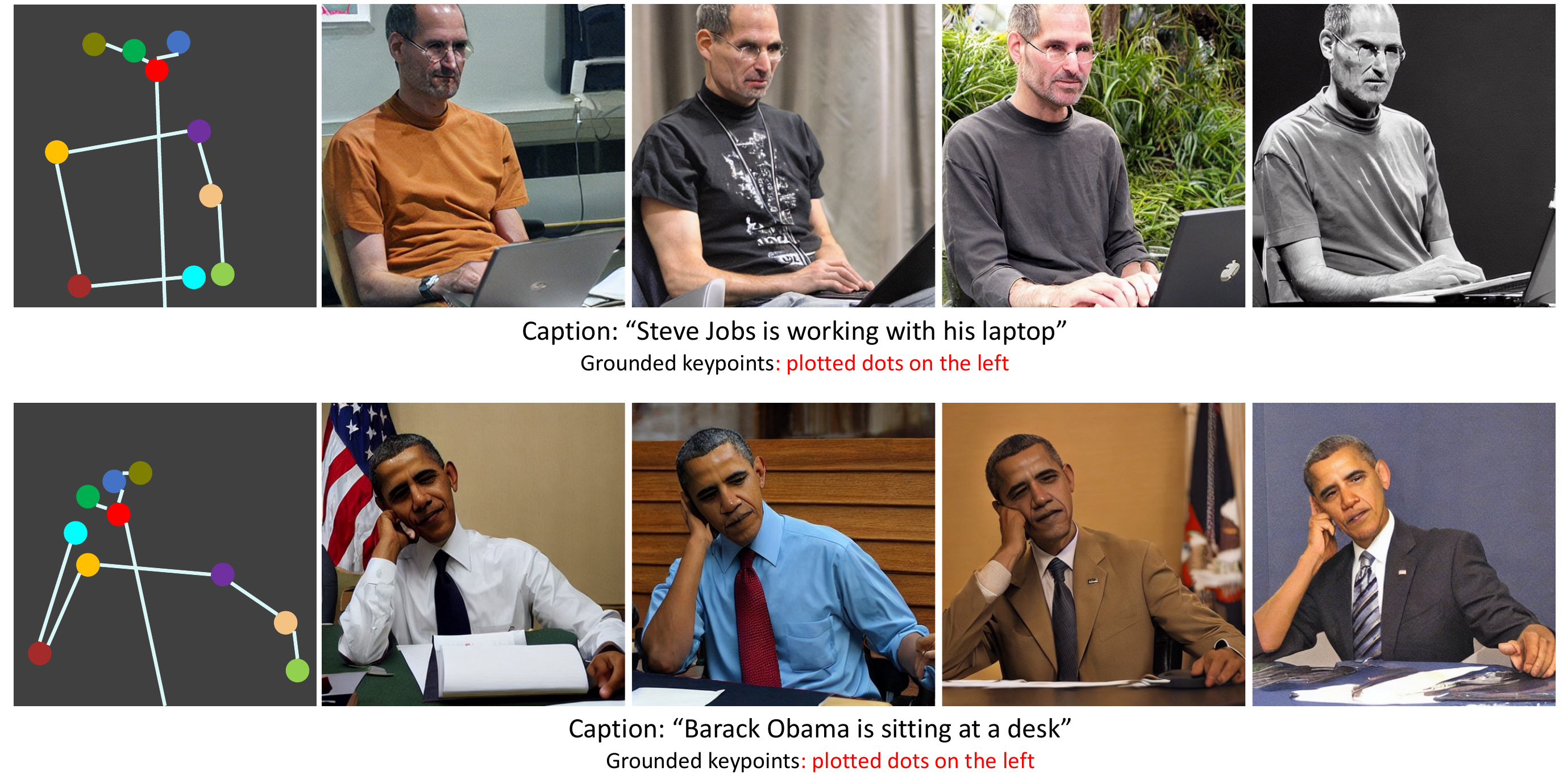}
    \caption{Results for keypoints grounded generation.}
    \label{fig:other_keypoint}
    \vspace{-0.1in}
\end{figure*}

\begin{figure*}[t]
    \centering
    \includegraphics[width=0.90\textwidth]{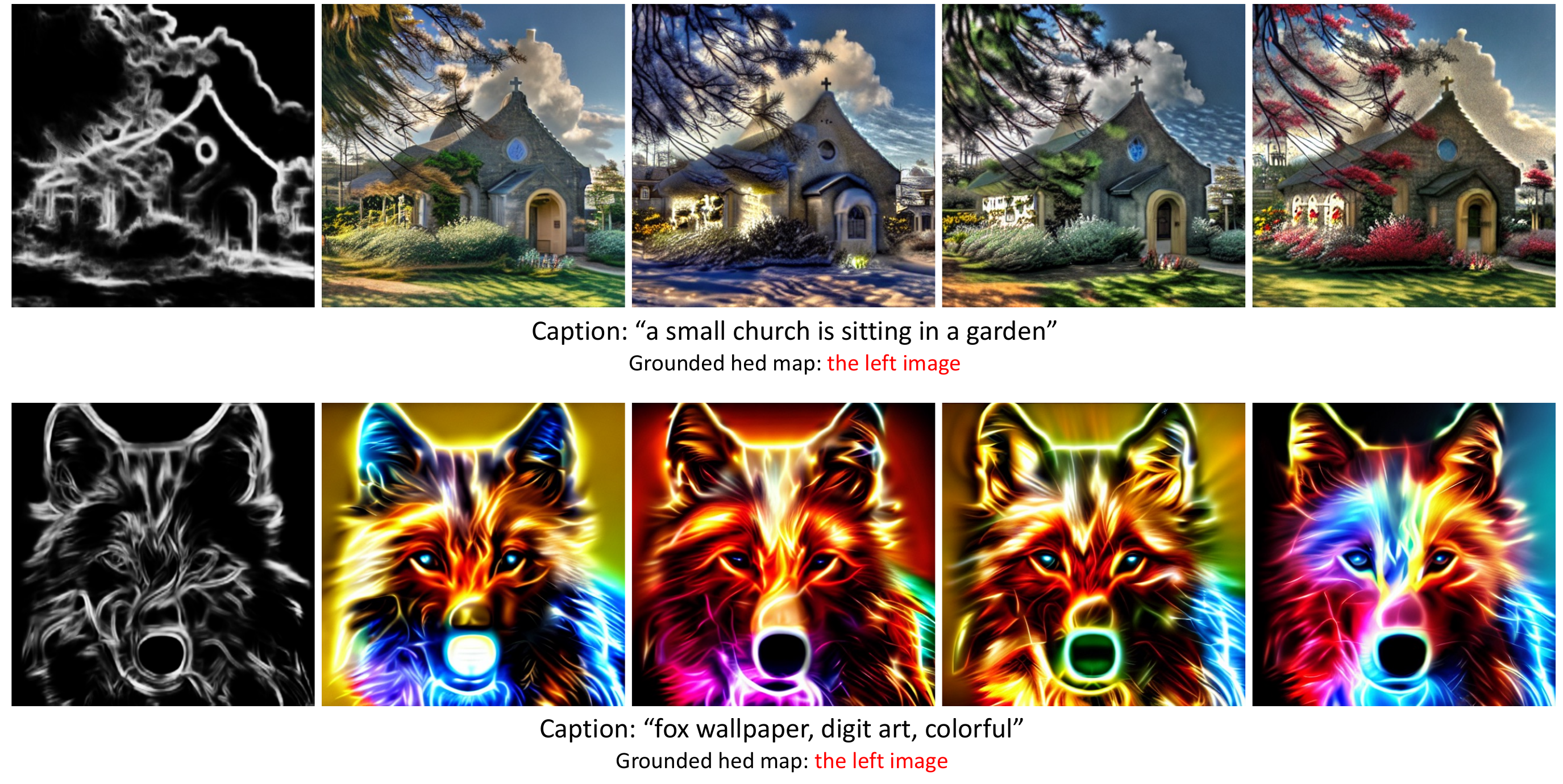}
    \caption{Results for HED map grounded generation.}
    \label{fig:other_hed}
    \vspace{-0.1in}
\end{figure*}

\begin{figure*}[t]
    \centering
    \includegraphics[width=0.90\textwidth]{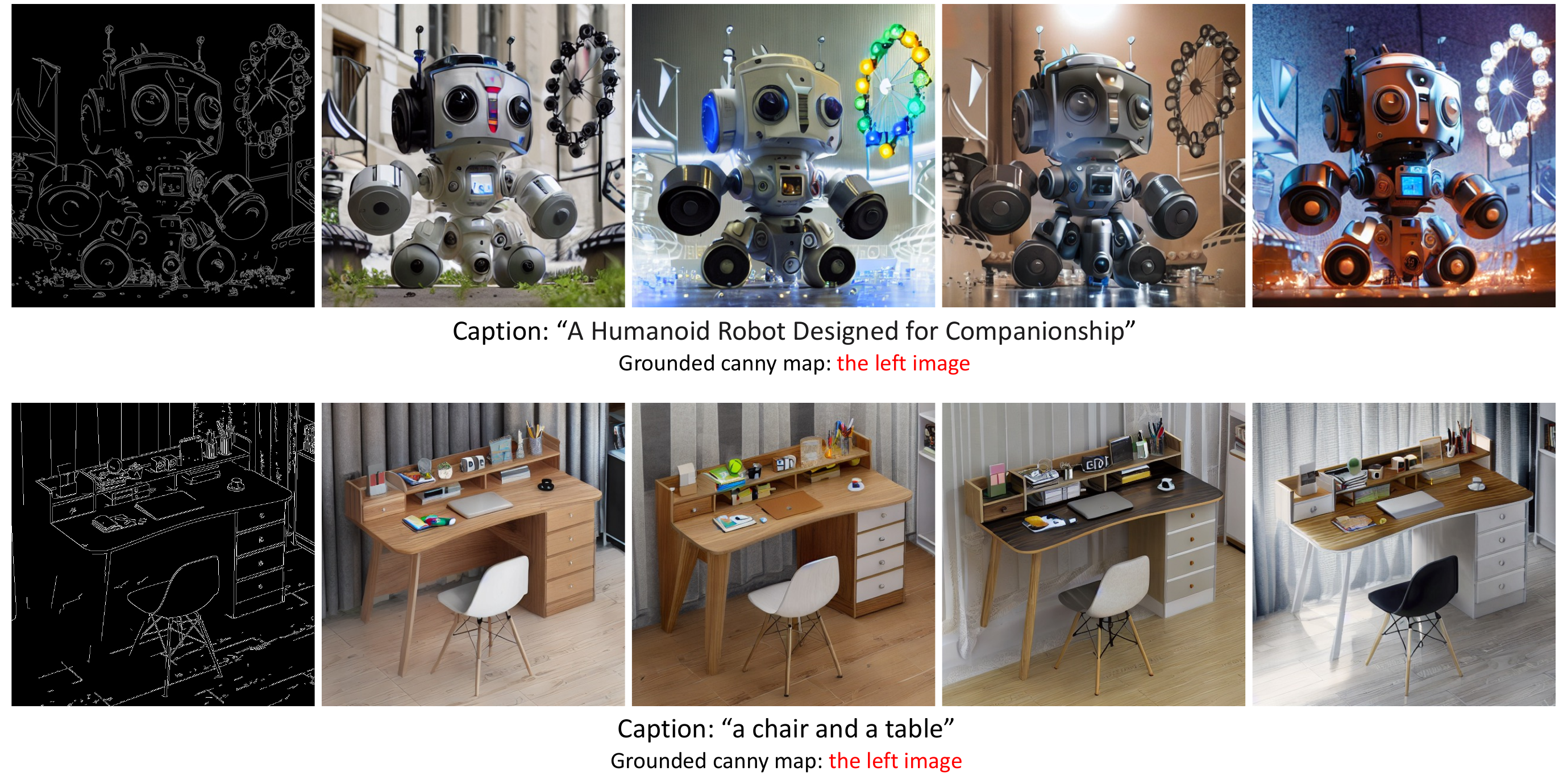}
    \caption{Results for canny map grounded generation.}
    \label{fig:other_canny}
    \vspace{-0.1in}
\end{figure*}

\begin{figure*}[t]
    \centering
    \includegraphics[width=0.90\textwidth]{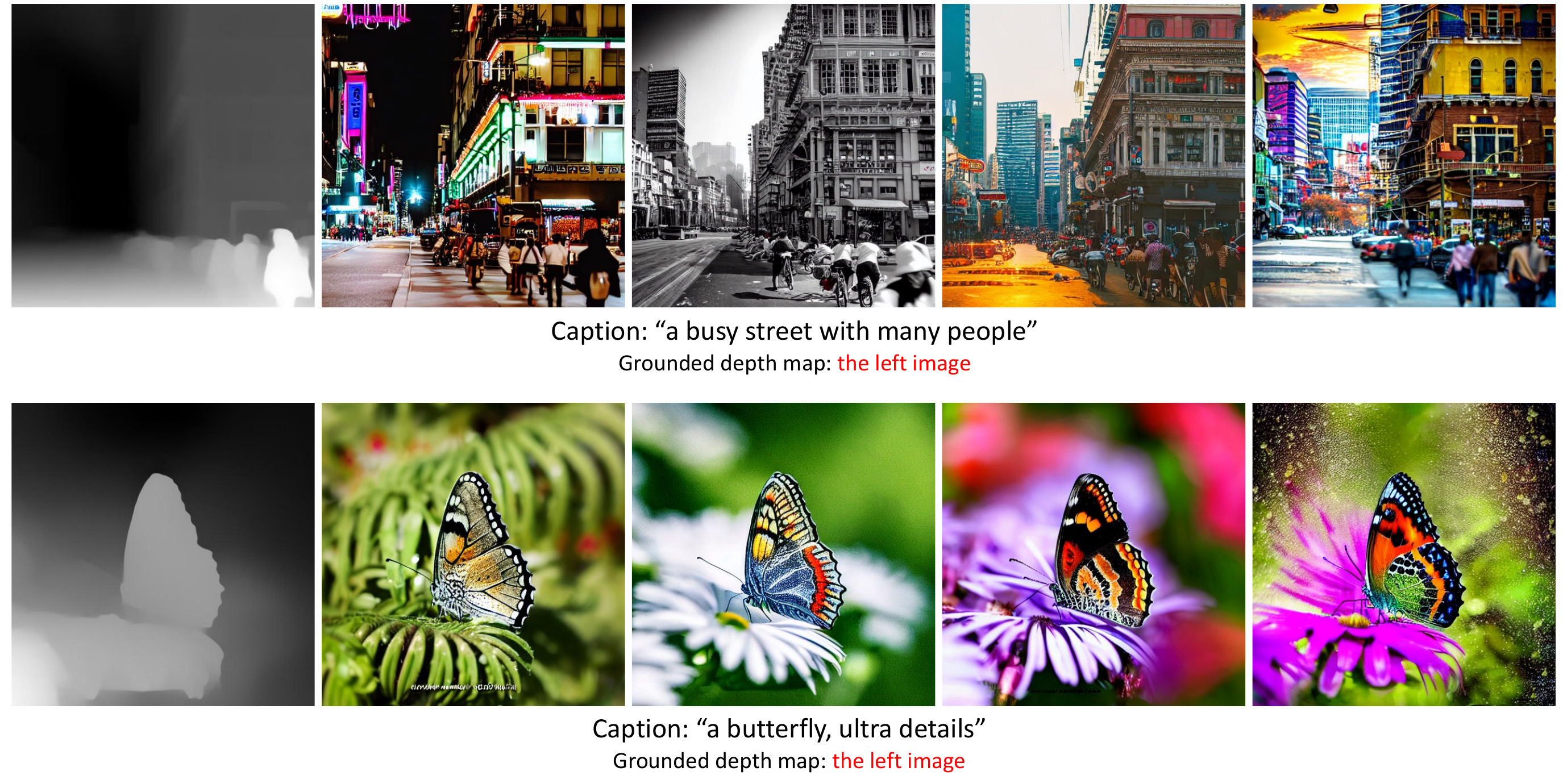}
    \caption{Results for depth map grounded generation.}
    \label{fig:other_depth}
    \vspace{-0.1in}
\end{figure*}

\begin{figure*}[t]
    \centering
    \includegraphics[width=0.90\textwidth]{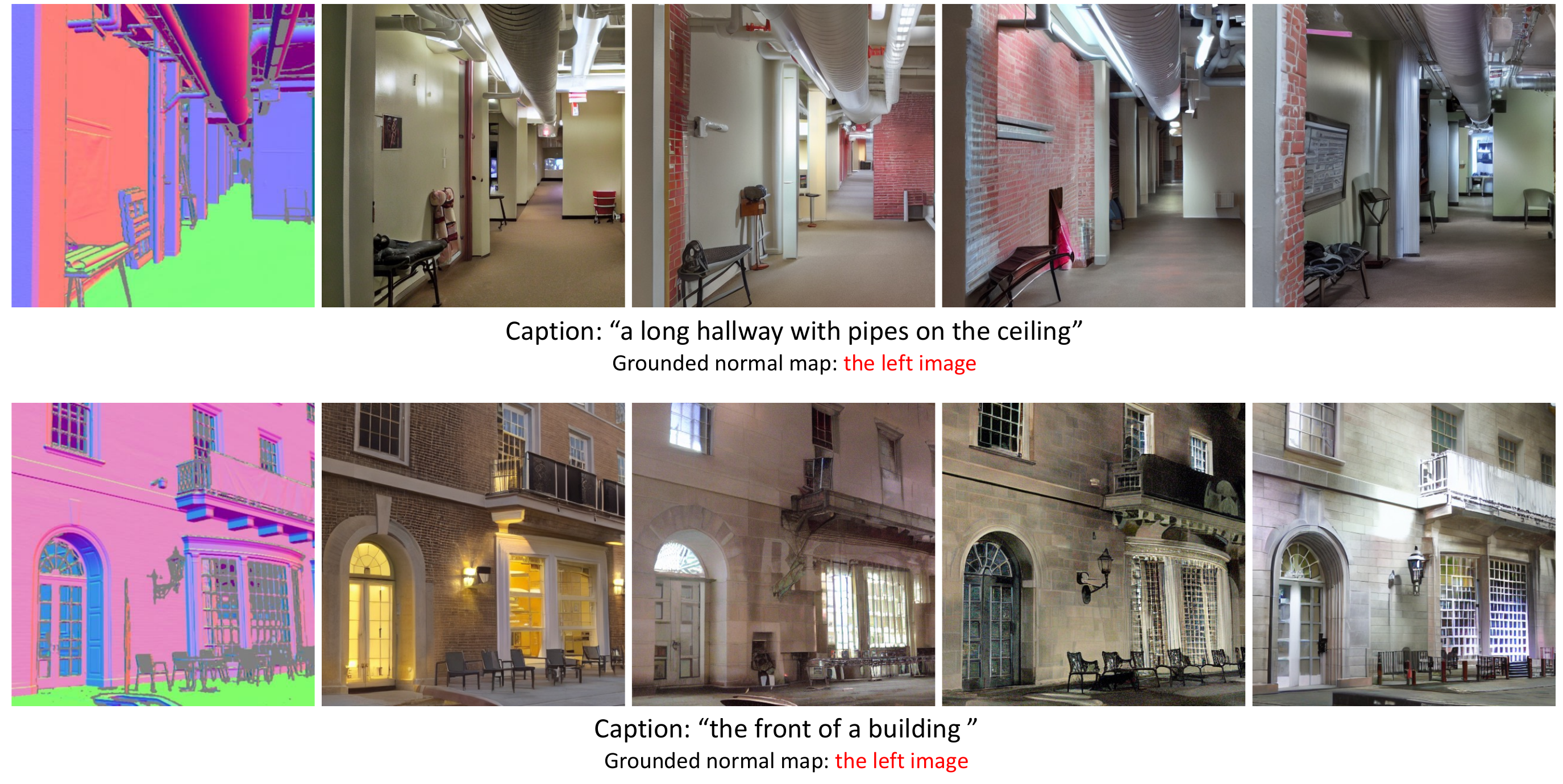}
    \caption{Results for normal map grounded generation.}
    \label{fig:other_normal}
    \vspace{-0.1in}
\end{figure*}

\begin{figure*}[t]
    \centering
    \includegraphics[width=0.90\textwidth]{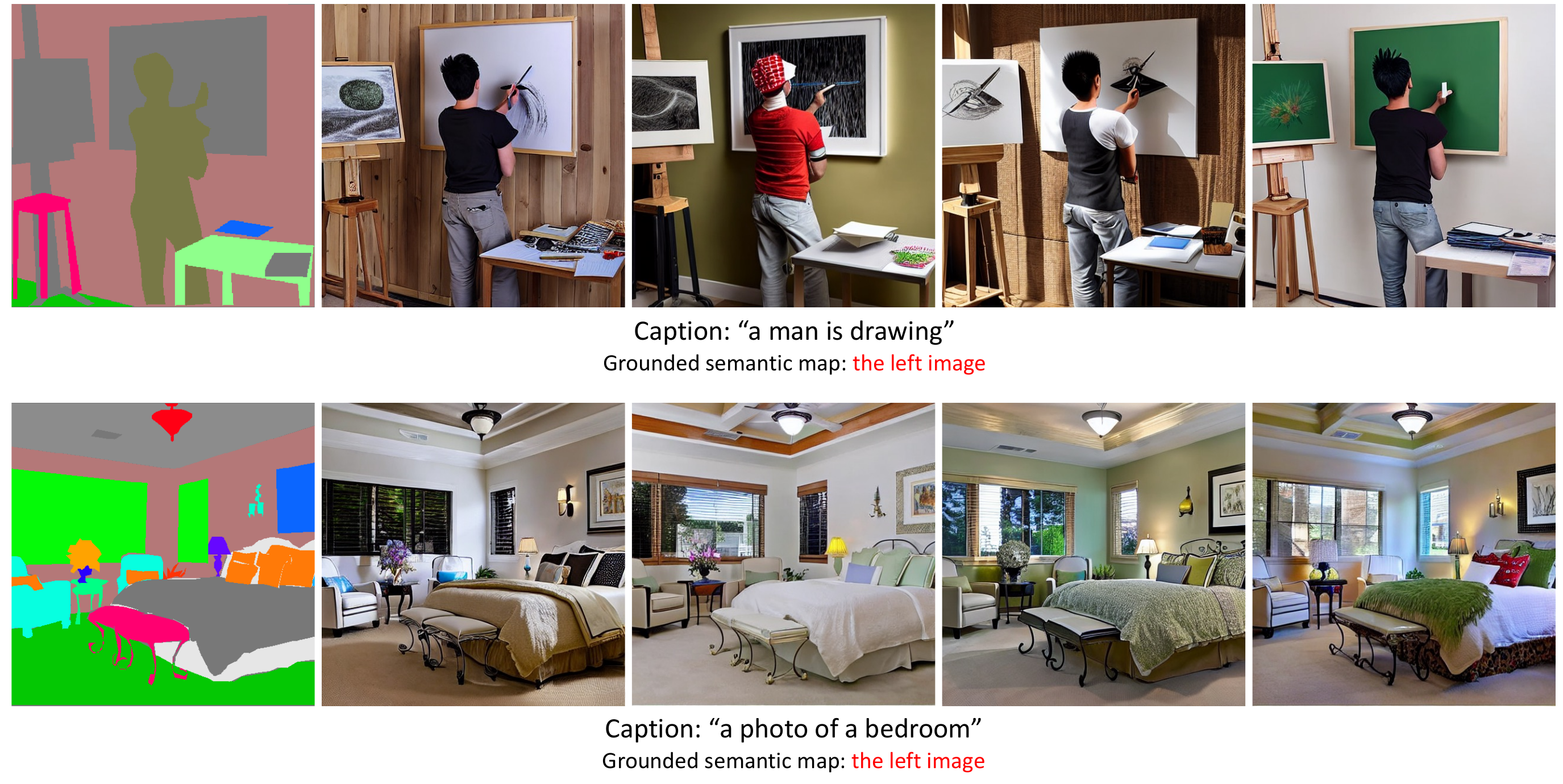}
    \caption{Results for semantic map grounded generation.}
    \label{fig:other_sem}
    \vspace{-0.1in}
\end{figure*}

Lastly, we show more grounded text2img results with bounding boxes in  Figure~\ref{fig:supp_groundtext2img} and other modality grounding results in  Figure~\ref{fig:other_keypoint}~\ref{fig:other_hed}~\ref{fig:other_canny}~\ref{fig:other_depth}~\ref{fig:other_normal}~\ref{fig:other_sem}. Note that our keypoint model only uses keypoint annotations from COCO~\cite{coco} which is not linked with person identity, but it can successfully utilize and combine the knowledge learned in the text2img training stage to control keypoints of a specific person. Out of curiosity, we also tested whether the keypoint grounding information learned on humans can be transferred to other non-humanoid categories such as cat or lamp for keypoint grounded generation, but we find that our model struggles in such cases even with scheduled sampling. Compared to bounding boxes, which only specify a coarse location and size of an object in the image and thus can be shared across all object categories, keypoints (i.e., object parts) are not always shareable across different categories. Thus, while keypoints enable more fine-grained control than boxes, they are less generalizable.

\end{document}